\documentclass[10pt,journal,compsoc]{IEEEtran}

\RequirePackage{snapshot}

\ifCLASSOPTIONcompsoc

  \usepackage[nocompress]{cite}
\else

  \usepackage{cite}
\fi

\ifCLASSINFOpdf
  \usepackage[pdftex]{graphicx}

\else

\fi

\usepackage{amsmath}
\usepackage{amssymb}

\usepackage{url}
\usepackage{hyperref}

\usepackage{multirow}

\def\eg{e.g.} 
\def\ie{i.e.} 
 
\def\etc{etc.}

\def\smallfig{\vspace{-0.35in}}

\usepackage{booktabs}
\usepackage{nicefrac}

\begin{document}
\title{Long-term Recurrent Convolutional Networks for \\
Visual Recognition and Description}

\author{
Jeff~Donahue,
Lisa~Anne~Hendricks,
Marcus~Rohrbach,
Subhashini~Venugopalan, 
Sergio~Guadarrama,
Kate~Saenko,
Trevor~Darrell
\IEEEcompsocitemizethanks{\IEEEcompsocthanksitem J. Donahue, L. A. Hendricks, M. Rohrbach, S. Guadarrama, and T. Darrell are with the Department of Electrical Engineering and Computer Science, UC Berkeley, Berkeley, CA.
\IEEEcompsocthanksitem M. Rohrbach and T. Darrell are additionally affiliated with the International Computer Science Institute, Berkeley, CA.
\IEEEcompsocthanksitem S. Venugopalan is with the Department of Computer Science, UT Austin, Austin, TX.
\IEEEcompsocthanksitem K. Saenko is with the Department of Computer Science, UMass Lowell, Lowell, MA.}
\thanks{Manuscript received November 30, 2015.}}

\IEEEtitleabstractindextext{
\begin{abstract}

Models based on deep convolutional networks have dominated recent image interpretation tasks; we investigate whether models which are also recurrent are effective for tasks involving sequences, visual and otherwise.
We describe a class of recurrent convolutional architectures which is end-to-end trainable and suitable for large-scale visual understanding tasks, and demonstrate the value of these models for activity recognition, image captioning, and video description.
In contrast to previous models which assume a fixed visual representation or perform simple temporal averaging for sequential processing, recurrent convolutional models are ``doubly deep'' in that they learn compositional representations in space and time.
Learning long-term dependencies is possible when nonlinearities are incorporated into the network state updates.
Differentiable recurrent models are appealing in that they can directly map variable-length inputs (\eg, videos)
to variable-length outputs (\eg, natural language text)
and can model complex temporal dynamics; yet they can be optimized with backpropagation.
Our recurrent sequence models are directly connected to modern visual convolutional network models and can be jointly trained to learn temporal dynamics and convolutional perceptual representations.
Our results show that such models have distinct advantages over state-of-the-art models for recognition or generation which are separately defined or optimized.

\end{abstract}

}

\maketitle

\IEEEdisplaynontitleabstractindextext

\IEEEpeerreviewmaketitle

\IEEEraisesectionheading{\section{Introduction}\label{sec:introduction}}

\begin{figure}[t!]
\begin{center}
\includegraphics[width=\linewidth]{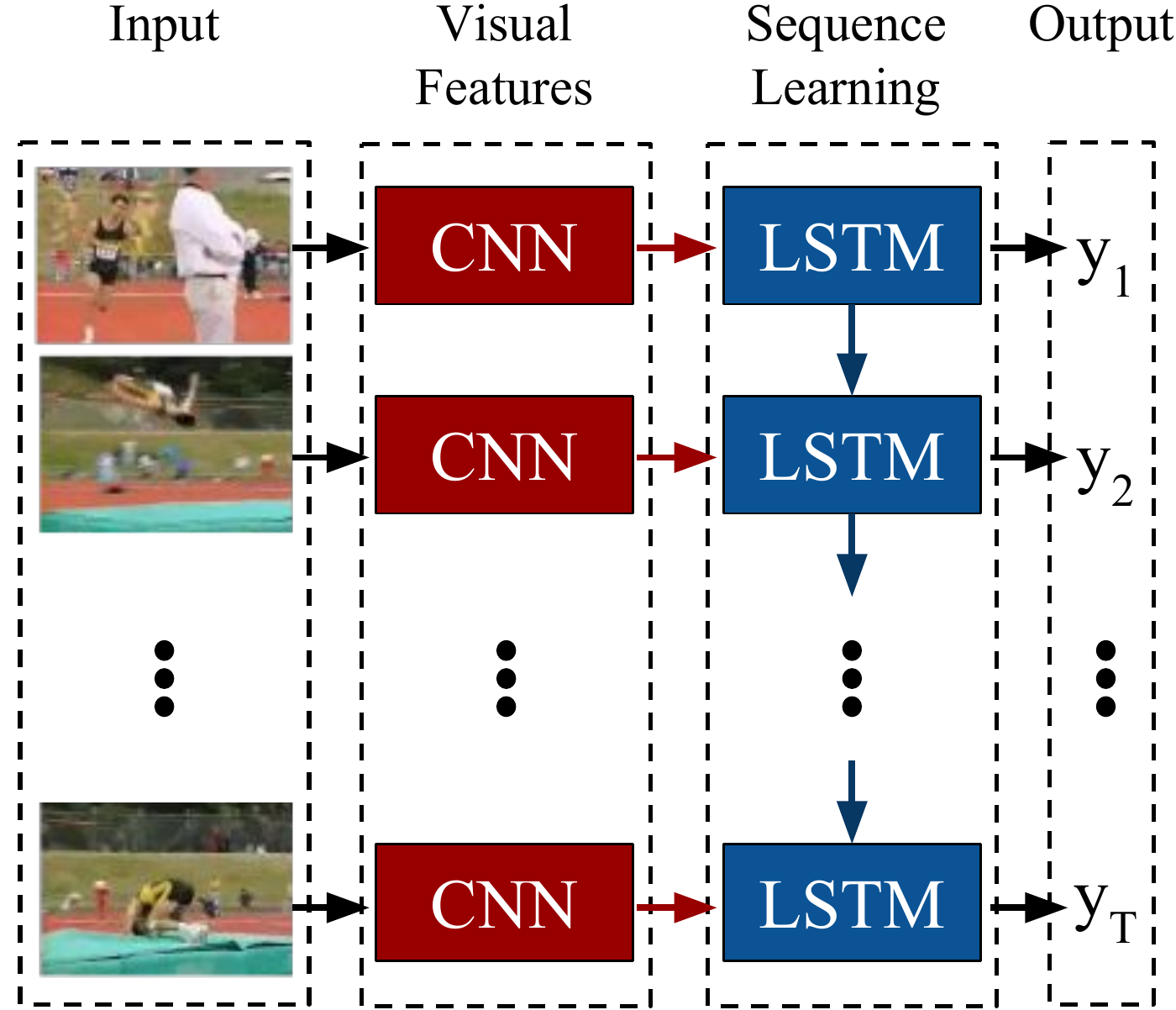}
\end{center}
 \caption{
We propose \textit{Long-term Recurrent Convolutional Networks} (LRCNs), a class of architectures leveraging the strengths of rapid progress in CNNs for visual recognition problems, and the growing desire to apply such models to time-varying inputs and outputs.
LRCN processes the (possibly) variable-length visual input (left) with a CNN (middle-left), whose outputs are fed into a stack of recurrent sequence models (\textit{LSTMs}, middle-right), which finally produce a variable-length prediction (right).
Both the CNN and LSTM weights are shared across time, resulting in a representation that scales to arbitrarily long sequences.
}
\label{fig:teaser}
\vspace{-0.2in}
\end{figure}

Recognition and description of images and videos is a fundamental challenge of computer vision.
Dramatic progress has been achieved by supervised convolutional neural network (CNN) models on image recognition tasks, and a number of extensions to process video have been recently proposed. Ideally, a video model should allow processing of variable length input sequences, and also provide for variable length outputs, including generation of full-length sentence descriptions that go beyond conventional one-versus-all prediction tasks.
In this paper we propose \textit{Long-term Recurrent Convolutional Networks} (LRCNs), a class of architectures for visual recognition and description which combines convolutional layers and long-range temporal recursion and is end-to-end trainable (Figure~\ref{fig:teaser}).
We instantiate our architecture for specific video activity recognition, image caption generation, and video description tasks as described below.

Research on CNN models for video processing has considered learning 3D spatio-temporal filters over raw sequence data \cite{Ji13tpami,baccouche2011sequential}, and learning of frame-to-frame representations which incorporate instantaneous optic flow or trajectory-based models aggregated over fixed windows or video shot segments~\cite{karpathy14cvpr,simonyan2014two}.
Such models explore two extrema of perceptual time-series representation learning: either learn a fully general time-varying weighting, or apply simple temporal pooling. 
Following the same inspiration that motivates current deep convolutional models, we advocate for video recognition and description models which are also deep over temporal dimensions; \ie, have temporal recurrence of latent variables.
Recurrent Neural Network (RNN) models are ``deep in time'' -- explicitly so when unrolled -- and form implicit compositional representations in the time domain.
Such ``deep'' models predated deep spatial convolution models in the literature~\cite{rumelhart1985learning,williams1989learning}.

The use of RNNs in perceptual applications has been explored for many decades, with varying results.
A significant limitation of simple RNN models which strictly integrate state information over time is known as the ``vanishing gradient'' effect: the ability to backpropagate an error signal through a long-range temporal interval becomes increasingly difficult in practice. 
\textit{Long Short-Term Memory} (LSTM) units, first proposed in \cite{hochreiter1997long}, are recurrent modules which enable long-range learning.
LSTM units have hidden state augmented with nonlinear mechanisms to allow state to propagate without modification, be updated, or be reset, using simple learned gating functions.
LSTMs have recently been demonstrated to be capable of large-scale learning of speech recognition~\cite{graves2014towards} and language translation models~\cite{sutskever14nips,cho2014properties}.

We show here that convolutional networks with recurrent units are generally applicable to visual time-series modeling, and argue that in visual tasks where static or flat temporal models have previously been employed, LSTM-style RNNs can provide significant improvement when ample training data are available to learn or refine the representation.
Specifically, we show that LSTM type models provide for improved recognition on conventional video activity challenges and enable a novel end-to-end optimizable mapping from image pixels to sentence-level natural language descriptions.
We also show that these models improve generation of descriptions from intermediate visual representations derived from conventional visual models.

We instantiate our proposed architecture in three experimental settings (Figure~\ref{fig:archs}).
First, we show that directly connecting a visual convolutional model to deep LSTM networks, we are able to train video recognition models that capture temporal state dependencies (Figure~\ref{fig:archs} left; Section~\ref{sec:activityRecognition}).
While existing labeled video activity datasets may not have actions or activities with particularly complex temporal dynamics, we nonetheless observe significant improvements on conventional benchmarks.

Second, we explore end-to-end trainable image to sentence mappings.
Strong results for machine translation tasks have recently been reported~\cite{sutskever14nips,cho2014properties}; such models are encoder-decoder pairs based on LSTM networks.
We propose a multimodal analog of this model, and describe an architecture which uses a visual convnet to encode a deep state vector, and an LSTM to decode the vector into a natural language string (Figure~\ref{fig:archs} middle; Section~\ref{sec:imageDescription}).
The resulting model can be trained end-to-end on large-scale image and text datasets, and even with modest training provides competitive generation results compared to existing methods.

Finally, we show that LSTM decoders can be driven directly from conventional computer vision methods which predict higher-level discriminative labels, such as the semantic video role tuple predictors in \cite{rohrbach13iccv} (Figure~\ref{fig:archs}, right; Section~\ref{sec:videoDescription}).
While not end-to-end trainable, such models offer architectural and performance advantages over previous statistical machine translation-based approaches.

We have realized a generic framework for recurrent models
in the widely adopted deep learning framework \textit{Caffe} \cite{jia2014caffe},
including ready-to-use implementations of RNN and LSTM units.
(See \url{http://jeffdonahue.com/lrcn/}.)

\section{Background: Recurrent Networks}
\label{sec:background}

\begin{figure}
\begin{center}
\includegraphics[width=0.96\linewidth]{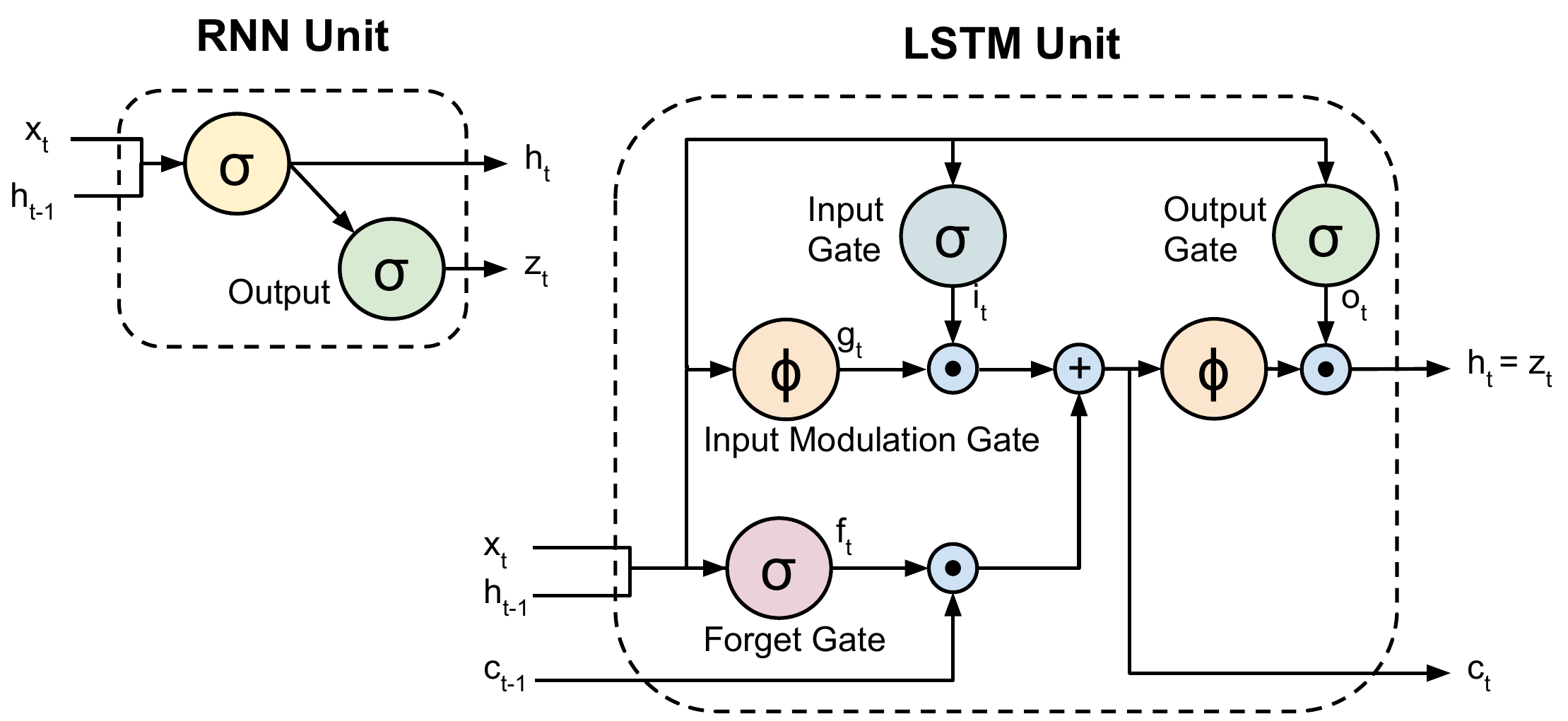}
\end{center}
\caption{
A diagram of a basic RNN cell (left) and an LSTM memory cell (right)
used in this paper (from~\cite{zaremba2014learning}, a slight simplification of the architecture described in
\cite{graves2013generating}, which was derived from the LSTM initially proposed in \cite{hochreiter1997long}).
}
\label{fig:lstm}
\end{figure}

Traditional recurrent neural networks (RNNs, Figure \ref{fig:lstm}, left) model temporal dynamics by mapping input sequences to hidden states, and hidden states to outputs via the following recurrence equations (Figure \ref{fig:lstm}, left):
\begin{align*}
h_t &= g(W_{xh}x_t + W_{hh}h_{t-1} + b_h) \\
z_t &= g(W_{hz}h_t + b_z)
\end{align*}
where $g$ is an element-wise non-linearity, such as a sigmoid or hyperbolic tangent, $x_t$ is the input,
$h_t \in \mathbb{R}^N$ is the hidden state with $N$ hidden units, and $z_t$ is the output at time $t$.
For a length $T$ input sequence $\langle x_1, x_2, ..., x_T \rangle$, the updates above are computed sequentially as $h_1$ (letting $h_0 = 0$), $z_1$, $h_2$, $z_2$, ..., $h_T$, $z_T$.

Though RNNs have proven successful on tasks such as speech recognition \cite{vinyals2012revisiting} and text generation \cite{sutskever2011generating}, it can be difficult to train them to learn long-term dynamics, likely due in part to the vanishing and exploding gradients problem~\cite{hochreiter1997long} that can result from propagating the gradients down through the many layers of the recurrent network, each corresponding to a particular time step.
LSTMs provide a solution by incorporating memory units that explicitly allow the network to learn when to ``forget'' previous hidden states and when to update hidden states given new information.
As research on LSTMs has progressed, hidden units with varying connections within the memory unit have been proposed.
We use the LSTM unit as described in \cite{zaremba2014learning} (Figure~\ref{fig:lstm}, right), a slight simplification of the one described in~\cite{graves2014towards}, which was derived from the original LSTM unit proposed in~\cite{hochreiter1997long}.
Letting $
\sigma(x) = \left(1 + e^{-x}\right)^{-1}
$ be the \textit{sigmoid} non-linearity which squashes real-valued inputs to a $[0, 1]$ range, and letting $
\tanh(x) = \frac{e^x - e^{-x}}{e^x + e^{-x}} = 2\sigma(2x) - 1
$ be the \textit{hyperbolic tangent} non-linearity, similarly squashing its inputs to a $[-1, 1]$ range,
the LSTM updates for time step $t$ given inputs $x_t$, $h_{t-1}$, and $c_{t-1}$ are:
\begin{align*}
i_t &= \sigma(W_{xi}x_t + W_{hi}h_{t-1} + b_i) \\
f_t &= \sigma(W_{xf}x_t + W_{hf}h_{t-1} + b_f) \\
o_t &= \sigma(W_{xo}x_t + W_{ho}h_{t-1} + b_o) \\
g_t &=  \tanh(W_{xc}x_t + W_{hc}h_{t-1} + b_c) \\
c_t &= f_t \odot c_{t-1} + i_t \odot g_t \\
h_t &= o_t \odot \tanh(c_t)
\end{align*}
$x \odot y$ denotes the element-wise product of vectors $x$ and $y$.

In addition to a hidden unit $h_t \in \mathbb{R}^N$, the LSTM includes an input gate $i_t \in \mathbb{R}^N$, forget gate $f_t \in \mathbb{R}^N$, output gate $o_t \in \mathbb{R}^N$, input modulation gate $g_t \in \mathbb{R}^N$, and memory cell $c_t \in \mathbb{R}^N$.
The memory cell unit $c_t$ is a sum of two terms: the previous memory cell unit $c_{t-1}$ which is modulated by $f_t$, and $g_t$, a function of the current input and previous hidden state, modulated by the input gate $i_t$.
Because $i_t$ and $f_t$ are sigmoidal, their values lie within the range $[0,1]$, and $i_t$ and $f_t$ can be thought of as knobs that the LSTM learns to selectively forget its previous memory or consider its current input.
Likewise, the output gate $o_t$ learns how much of the memory cell to transfer to the hidden state.
These additional cells seem to enable the LSTM to learn complex and long-term temporal dynamics for a wide variety of sequence learning and prediction tasks.
Additional depth can be added to LSTMs by stacking them on top of each other, using the hidden state $h_t^{(\ell-1)}$ of the LSTM in layer $\ell-1$ as the input to the LSTM in layer $\ell$.

Recently, LSTMs have achieved impressive results on language tasks such as speech recognition \cite{graves2014towards} and machine translation  \cite{sutskever14nips,cho2014properties}.
Analogous to CNNs, LSTMs are attractive because they allow end-to-end fine-tuning.
For example, \cite{graves2014towards} eliminates the need for complex multi-step pipelines in speech recognition by training a deep bidirectional LSTM which maps spectrogram inputs to text.
Even with no language model or pronunciation dictionary, the model produces convincing text translations.
\cite{sutskever14nips} and \cite{cho2014properties} translate sentences from English to French with a multi-layer LSTM encoder and decoder.
Sentences in the source language are mapped to a hidden state using an encoding LSTM, and then a decoding LSTM maps the hidden state to a sequence in the target language.
Such an encoder-decoder scheme allows an input sequence of arbitrary length to be mapped to an output sequence of different length.
The sequence-to-sequence architecture for machine translation circumvents the need for language models.

The advantages of LSTMs for modeling sequential data in vision problems are twofold.
First, when integrated with current vision systems, LSTM models are straightforward to fine-tune end-to-end.
Second, LSTMs are not confined to fixed length inputs or outputs allowing simple modeling for sequential data of varying lengths, such as text or video.
We next describe a unified framework to combine recurrent models such as LSTMs with deep convolutional networks to form end-to-end trainable networks capable of complex visual and sequence prediction tasks.

\section{Long-term Recurrent Convolutional Network (LRCN) model}
\label{sec:model}
This work proposes a Long-term Recurrent Convolutional Network (LRCN) model combining a deep hierarchical visual feature extractor (such as a CNN) with a model that can learn to recognize and synthesize temporal dynamics for tasks involving sequential data (inputs or outputs), visual, linguistic, or otherwise.
Figure~\ref{fig:teaser} depicts the core of our approach.
LRCN works by passing each visual input $x_t$ (an image in isolation, or a frame from a video) through a feature transformation $\phi_V(.)$ with parameters $V$,
usually a CNN, to produce a fixed-length vector representation $\phi_V(x_t)$.
The outputs of $\phi_V$ are then passed into a recurrent sequence learning module.

In its most general form, a recurrent model has parameters $W$,
and maps an input $x_t$ and a previous time step hidden state $h_{t-1}$
to an output $z_t$ and updated hidden state $h_t$.
Therefore, inference must be run sequentially
(\ie, from top to bottom, in the \textit{Sequence Learning} box of Figure~\ref{fig:teaser}),
by computing in order:
$h_1 = f_W(x_1, h_0) = f_W(x_1, 0)$, then $
h_2 = f_W(x_2, h_1)$, \etc, up to $h_T$.
Some of our models stack multiple LSTMs atop one another as described in Section~\ref{sec:background}.

To predict a distribution $P(y_t)$ over outcomes $y_t \in C$ (where $C$ is a discrete, finite set of outcomes) at time step $t$, the outputs $z_t \in \mathbb{R}^{d_z}$ of the sequential model are passed through a linear prediction layer $\hat{y}_t = W_z z_t + b_z$, where $W_z \in \mathbb{R}^{|C| \times d_z}$ and $b_z \in \mathbb{R}^{|C|}$ are learned parameters.
Finally, the predicted distribution $P(y_t)$ is computed by taking the softmax of $\hat{y}_t$:
$
P(y_t = c) = \mathrm{softmax}(\hat{y}_t) =
\frac{\exp(\hat{y}_{t,c})}{\sum\limits_{c' \in C} \exp(\hat{y}_{t,c'})}
$.

The success of recent deep models for object recognition~\cite{krizhevsky2012imagenet,vggnet,szegedy2014going} suggests that strategically composing many ``layers'' of non-linear functions can result in powerful models for perceptual problems.
For large $T$, the above recurrence indicates that the last few predictions from a recurrent network with $T$ time steps are computed by a very ``deep'' ($T$ layer) non-linear function, suggesting that the resulting recurrent model may have similar representational power to a $T$ layer deep network.
Critically, however, the sequence model's weights $W$ are reused at every time step, forcing the model to learn generic time step-to-time step dynamics (as opposed to dynamics conditioned on $t$, the sequence index) and preventing the parameter size from growing in proportion to the maximum sequence length.

In most of our experiments, the visual feature transformation $\phi$ corresponds to the activations in some layer of a deep CNN.
Using a visual transformation $\phi_V(.)$ which is time-invariant and independent at each time step has the important advantage of making the expensive convolutional inference and training parallelizable over all time steps of the input, facilitating the use of fast contemporary CNN implementations whose efficiency relies on independent batch processing, and end-to-end optimization of the visual and sequential model parameters $V$ and $W$.

\begin{figure*}[t]
\center
\includegraphics[width=.9\linewidth]{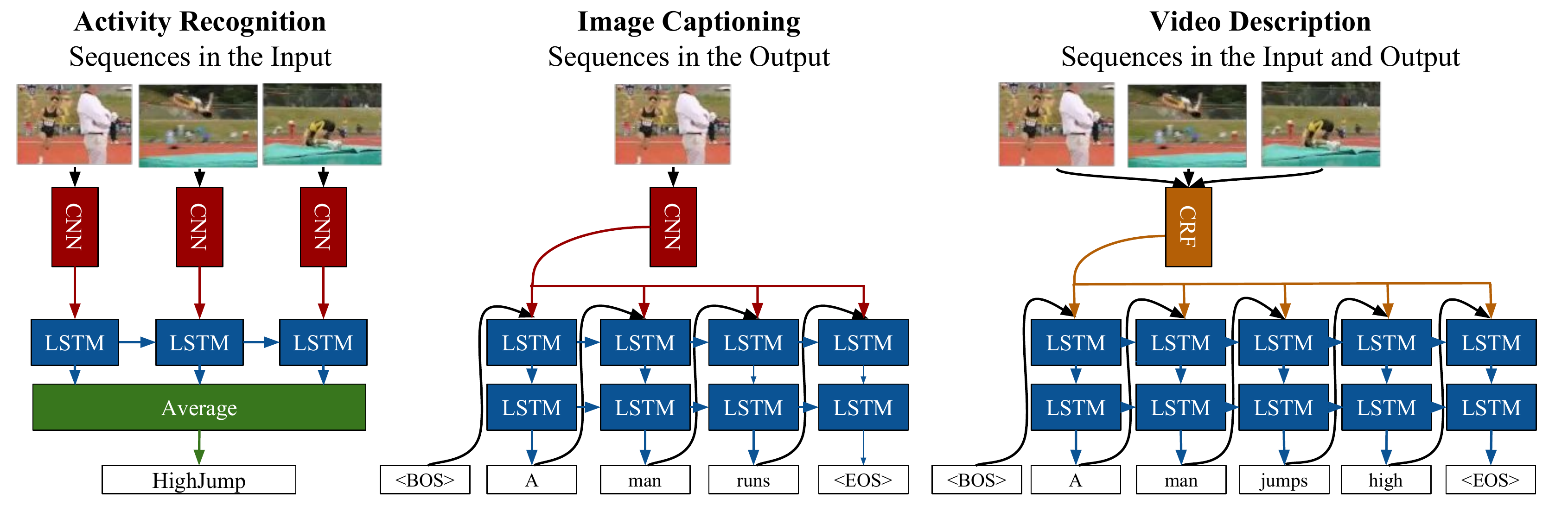}
\vspace{-0.5cm}
  \caption{Task-specific instantiations of our LRCN model for activity recognition, image description, and video description.}
\label{fig:archs}
\end{figure*}

We consider three vision problems (activity recognition, image description and video description), each of which instantiates one of the following broad classes of sequential learning tasks:

\begin{enumerate}
\item  Sequential input, static output (Figure \ref{fig:archs}, left):
$
\langle
x_1, x_2, ..., x_T
\rangle
\mapsto
y
$.
The visual activity recognition problem can fall under this umbrella, with videos of arbitrary length $T$ as input, but with the goal of predicting a single label like \textit{running} or \textit{jumping} drawn from a fixed vocabulary.

\item  Static input, sequential output (Figure \ref{fig:archs}, middle):
$
x
\mapsto
\langle
y_1, y_2, ..., y_T
\rangle
$.
The image captioning problem fits in this category, with a static (non-time-varying) image as input, but a much larger and richer label space consisting of \textit{sentences} of any length.

\item  Sequential input and output (Figure \ref{fig:archs}, right):
$
\langle
x_1, x_2, ..., x_T
\rangle
\mapsto
\langle
y_1, y_2, ..., y_{T'}
\rangle
$.
In tasks such as video description, both the visual input and output are time-varying,
and in general the number of input and output time steps may differ (\ie, we may have $T \ne T'$).
In video description, for example, the number of frames in the video should not constrain the length of
(number of words in) the natural language description.
\end{enumerate}

In the previously described generic formulation of recurrent models,
each instance has $T$ inputs $\langle x_1, x_2, ..., x_T \rangle$
and $T$ outputs $\langle y_1, y_2, ..., y_T \rangle$.
Note that this formulation does not align cleanly with any of the three problem classes described above -- in the first two classes, either the input or output is static, and in the third class, the input length $T$ need not match the output length $T'$.
Hence, we describe how we adapt this formulation in our hybrid model
to each of the above three problem settings.

With sequential inputs and static outputs (class 1),
we take a late-fusion approach to merging the per-time step predictions $\langle y_1, y_2, ..., y_T \rangle$
into a single prediction $y$ for the full sequence.
With static inputs $x$ and sequential outputs (class 2),
we simply duplicate the input $x$ at all $T$ time steps: $
\forall t \in \{1, 2, ..., T\}: x_t := x
$.
Finally, for a sequence-to-sequence problem with (in general) different input and output lengths (class 3),
we take an ``encoder-decoder'' approach,
as proposed for machine translation by~\cite{cho14emnlp,sutskever14nips}.
In this approach, one sequence model, the \textit{encoder}, maps the input sequence to a fixed-length vector,
and another sequence model, the \textit{decoder}, unrolls this vector to a sequential output of arbitrary length.
Under this type of model, a run of the full system on one instance occurs over $T + T' - 1$ time steps.
For the first $T$ time steps, the encoder processes the input $x_1, x_2, ..., x_T$,
and the decoder is inactive until time step $T$,
when the encoder's output is passed to the decoder,
which in turn predicts the first output $y_1$.
For the latter $T'-1$ time steps,
the decoder predicts the remainder of the output $y_2, y_3, ..., y_{T'}$
with the encoder inactive.
This encoder-decoder approach, as applied to the video description task,
is depicted in Section~\ref{sec:videoDescription},
Figure~\ref{fig:videoDescription:appraoch} (left).

Under the proposed system, the parameters $\left(V, W\right)$
of the model's visual and sequential components can be jointly optimized
by maximizing the likelihood of the ground truth outputs $y_t$ at each time step $t$,
conditioned on the input data and labels up to that point $\left(x_{1:t}, y_{1:t-1}\right)$.
In particular, for a training set $\mathcal{D}$ of labeled sequences $(x_t, y_t)_{t=1}^T \in \mathcal{D}$,
we optimize parameters $\left(V, W\right)$ to minimize the expected negative log likelihood of a sequence sampled from the training set $
\mathcal{L}(V, W, \mathcal{D}) =
-
\frac{1}{|\mathcal{D}|}
\sum_{(x_t, y_t)_{t=1}^T \in \mathcal{D}}
\sum_{t=1}^{T}
\log
P(
y_t |
x_{1:t},
y_{1:t-1},
V, W)
$.

One of the most appealing aspects of the described system is the ability to learn the parameters ``end-to-end,'' such that the parameters $V$ of the visual feature extractor learn to pick out the aspects of the visual input that are relevant to the sequential classification problem.
We train our LRCN models using stochastic gradient descent, with backpropagation used to compute the gradient $
\nabla_{V,W} \mathcal{L}(V, W, \tilde{\mathcal{D}})
$
of the objective $\mathcal{L}$ with respect to all parameters $(V, W)$ over minibatches $\tilde{\mathcal{D}} \subset \mathcal{D}$ sampled from the training dataset $\mathcal{D}$.

We next demonstrate the power of end-to-end trainable hybrid convolutional and recurrent networks by exploring three applications: activity recognition, image captioning, and video description.

\section{Activity recognition}
\label{sec:activityRecognition}
Activity recognition is an instance of the first class of sequential learning tasks described above:
each frame in a length $T$ sequence is the input to a single convolutional network
(\ie, the convnet weights are tied across time).
We consider both RGB and flow as inputs to our recognition system.  
Flow is computed with \cite{brox2004high} and transformed into a ``flow image'' by scaling and shifting $x$ and $y$ flow values to a range of $[-128, +128]$.
A third channel for the flow image is created by calculating the flow magnitude.

During training, videos are resized to $240\times320$ and we augment our data by using $227\times227$ crops and mirroring.
Additionally, we train the LRCN networks with video clips of 16 frames, even though the UCF101 videos are generally much longer (on the order of 100 frames when extracting frames at 30 FPS). 
Training on shorter video clips can be seen as analogous to training on image crops and is a useful method of data augmentation.
LRCN is trained to predict the video's activity class at each time step.
To produce a single label prediction for an entire video clip, we average the label probabilities -- the outputs of the network's softmax layer -- across all frames and choose the most probable label.
At test time, we extract 16 frame clips with a stride of 8 frames from each video and average across all clips from a single video.

The CNN base of LRCN in our activity recognition experiments
is a hybrid of the \textit{CaffeNet}~\cite{jia2014caffe} reference model
(a minor variant of \textit{AlexNet}~\cite{krizhevsky2012imagenet})
and the network used by Zeiler \& Fergus~\cite{zeiler2014visualizing}.
The network is pre-trained on the 1.2M image ILSVRC-2012~\cite{ILSVRC15} classification training subset of the ImageNet~\cite{deng09cvpr} dataset, giving the network a strong initialization to facilitate faster training and avoid overfitting to the relatively small video activity recognition datasets.
When classifying center crops, the top-1 classification accuracy is 60.2\% and 57.4\% for the hybrid and \textit{CaffeNet} reference models, respectively.

We compare LRCN to a single frame baseline model.
In our baseline model, $T$ video frames are individually classified by a CNN. 
As in the LSTM model, whole video classification is done by averaging scores across all video frames.

\subsection{Evaluation}

We evaluate our architecture on the UCF101 dataset \cite{soomro2012ucf101} which consists of over 12,000 videos categorized into 101 human action classes.  
The dataset is split into three splits, with just under 8,000 videos in the training set for each split.  

We explore various hyperparameters for the LRCN activity recognition architecture.
To explore different variants, we divide the first training split of UCF101 into a smaller training set ($\approx$6,000 videos) and a validation set ($\approx$3,000 videos).
We find that the most influential hyperparameters include the number of hidden units in the LSTM and whether $fc_6$ or $fc_7$ features are used as input to the LSTM.
We compare networks with $256$, $512$, and $1024$ LSTM hidden units.
When using flow as an input, more hidden units leads to better peformance with 1024 hidden units yielding a 1.7\% boost in accuracy in comparison to a network with 256 hidden units on our validation set.
In contrast, for networks with RGB input, the number of hidden units has little impact on the performance of the model.
We thus use 1024 hidden units for flow inputs, and 256 for RGB inputs.
We find that using $fc_6$ as opposed to $fc_7$ features improves accuracy when using flow as input on our validation set by 1\%.  
When using RGB images as input, the difference between using $fc_6$ or $fc_7$ features is quite small; using $fc_6$ features only increases accuracy by 0.2\%.
Because both models perform better with $fc_6$ features, we train our final models using $fc_6$ features (denoted by LRCN-${fc_6}$).
We also considered subsampling the frames input to the LSTM, but found that this hurts performance compared with using all frames.
Additionally, when training the LRCN network end-to-end, we found that aggressive dropout ($0.9$) was needed to avoid overfitting.

Table \ref{tab:average_accuracies} reports the average accuracy across the three standard test splits of UCF101.
Columns 2-3, compare video classification of LRCN against the baseline single frame architecture for both RGB and flow inputs.  
LRCN yields the best results for both RGB and flow and improves upon the baseline network by 0.83\% and 2.91\% respectively.
RGB and flow networks can be combined by computing a weighted average of network scores as proposed in \cite{simonyan2014two}.  
Like \cite{simonyan2014two}, we report two weighted averages of the predictions from the RGB and flow networks in Table~\ref{tab:average_accuracies} (right). 
Since the flow network outperforms the RGB network, weighting the flow network higher unsurprisingly leads to better accuracy.
In this case, LRCN outperforms the baseline single-frame model by 3.40\%.

\newcommand{\ucflbl}[1]{\textit{#1}}
Table~\ref{tab:class_accuracies} compares LRCN's accuracy with the single frame baseline model for individual classes on Split 1 of UCF101.
For the majority of classes, LRCN improves performance over the single frame model.  
Though LRCN performs worse on some classes including \ucflbl{Knitting} and \ucflbl{Mixing}, in general when LRCN performs worse, the loss in accuracy is not as substantial as the gain in accuracy for classes like \ucflbl{BoxingPunchingBag} and \ucflbl{HighJump}.
Consequently, accuracy is higher overall.

Table~\ref{tab:RGBvsFlow} compares accuracies for the LRCN flow and LRCN RGB models for individual classes on Split 1 of UCF101.
Note that for some classes the LRCN flow model outperforms the LRCN RGB model and vice versa.
One explanation is that activities which are better classified by the LRCN RGB model are best determined by which objects are present in the scene, while activities which are better classified by the LRCN flow model are best classified by the kind of motion in the scene.
For example, activity classes like \ucflbl{Typing} are highly correlated with the presence of certain objects, such as a keyboard, and are thus best learned by the LRCN RGB model.
Other activities such as \ucflbl{SoccerJuggling} include more generic objects which are frequently seen in other activities (soccer balls, people) and are thus best identified from class-specific motion cues.
Because RGB and flow signals are complementary, the best models take both into account.

LRCN shows clear improvement over the baseline single-frame system and is comparable to accuracy achieved by other deep models.
\cite{simonyan2014two} report the results on UCF101 by computing a weighted average between flow and RGB networks and achieve 87.6\%.
 \cite{karpathy14cvpr} reports 65.4\% accuracy on UCF101, which is substantially lower than LRCN.

\begin{table}
\begin{center}
\begin{tabular}{l|cc|cc}
\toprule
&\multicolumn{2}{|c}{\bf{Single Input Type}} & \multicolumn{2}{|c}{\bf{Weighted Average}}\\
\bf{Model} & RGB & Flow & $\nicefrac{1}{2}, \nicefrac{1}{2}$ &  $\nicefrac{1}{3}, \nicefrac{2}{3}$ \\ \hline
Single frame &  67.37  & 74.37 & 75.46    & 78.94\\
LRCN-fc$_6$ & \bf{68.20} & \bf{77.28} & 80.90 & \bf{82.34}\\
\bottomrule
\end{tabular}
\end{center}
\caption{
 Activity recognition: Comparing single frame models to LRCN networks for activity recognition on the UCF101~\cite{soomro2012ucf101} dataset, with RGB and flow inputs.
Average values across all three splits are shown.
LRCN consistently and strongly outperforms a model based on predictions from the underlying convolutional network architecture alone.
}
\label{tab:average_accuracies}
\vspace{-0.2in}
\end{table}

\begin{table}[btp]
\begin{center}
\begin{tabular}{lr|lr}
\toprule
Label & $\Delta$ & Label & $\Delta$ \\ \hline
BoxingPunchingBag & 40.82 & BoxingSpeedBag & -16.22\\
HighJump & 29.73 & Mixing & -15.56\\
JumpRope & 28.95 & Knitting & -14.71\\
CricketShot & 28.57 & Typing & -13.95\\
Basketball & 28.57 & Skiing & -12.50\\
WallPushups & 25.71 & BaseballPitch & -11.63\\
Nunchucks & 22.86 & BrushingTeeth & -11.11\\
ApplyEyeMakeup & 22.73 & Skijet & -10.71\\
HeadMassage & 21.95 & Haircut & -9.10\\
Drumming & 17.78 & TennisSwing & -8.16\\
\bottomrule
\end{tabular}
\end{center}
\caption{Activity recognition: comparison of improvement $\Delta$ in LRCN's per-class recognition accuracy versus the single-frame baseline.  Here we report results on all three splits of UCF101 (only results on the first split were presented in the paper).
$\Delta$ is the difference between LRCN's accuracy and the single-frame model's accuracy.}
\label{tab:class_accuracies}
\vspace{-0.38in}
\end{table}

\begin{table}[btp]
\begin{center}
\begin{tabular}{lr|lr}
\toprule
Label & $\Delta$ & Label & $\Delta$\\ \hline
BoxingPunchingBag & 57.14 & Typing & -44.19\\
PushUps & 53.33 &TennisSwing & -42.86\\
JumpRope & 50.00 & FieldHockeyPenalty &  -32.50\\
SoccerJuggling & 48.72 & BrushingTeeth  & -30.56\\
HandstandWalking & 44.12 & CuttingInKitchen & -30.30\\
Basketball & 40.00 & Skijet & -28.57\\
BodyWeightSquats & 38.46 & Mixing & -26.67\\
Lunges & 37.84 & Skiing &-25.00\\
Nunchucks & 34.29 & Knitting & -20.59\\
WallPushups & 34.29 & FloorGymnastics & -19.44\\
\bottomrule
\end{tabular}
\end{center}
\caption{Activity recognition: comparison of per-class recognition accuracy between the flow and RGB LRCN models. $\Delta$ is the difference between LRCN flow accuracy and LRCN RGB accuracy.}
\label{tab:RGBvsFlow}
\smallfig
\end{table}

\section{Image captioning}
\label{sec:imageDescription}
In contrast to activity recognition, the static image captioning task requires only a single invocation of a convolutional network since the input consists of a single image.
At each time step, both the image features and the previous word are provided as inputs to the sequence model, in this case
a stack of LSTMs (each with 1000 hidden units), which is used to learn the dynamics of the time-varying output sequence, natural language.

At time step $t$, the input to the bottom-most LSTM is the embedded word from the previous time step $y_{t-1}$.
Input words are encoded as ``one-hot'' vectors: vectors $
y \in \mathbb{R}^{K}
$ with a single non-zero component $y_i = 1$ denoting the $i^{\mathrm{th}}$ word in the vocabulary, where $K$ is the number of words in the vocabulary, plus one additional entry for the \texttt{<BOS>} (beginning of sequence) token which is always taken as $y_0$, the ``previous word'' at the first time step ($t=1$).
These one-hot vectors are then projected into an embedding space with dimension $d_e$ by multiplication $W_e y_t$ with a learned parameter matrix $W_e \in \mathbb{R}^{d_e \times K}$.
The result of a matrix-vector multiplication with a one-hot vector is the column of the matrix corresponding to the index of the single non-zero component of the one-hot vector.
$W_e$ can therefore be thought of as a ``lookup table,'' mapping each of the $K$ words in the vocabulary to a $d_e$-dimensional vector.

\begin{figure*}
\centering
\begin{tabular}{c|c|c}
\includegraphics[width=0.25\textwidth]{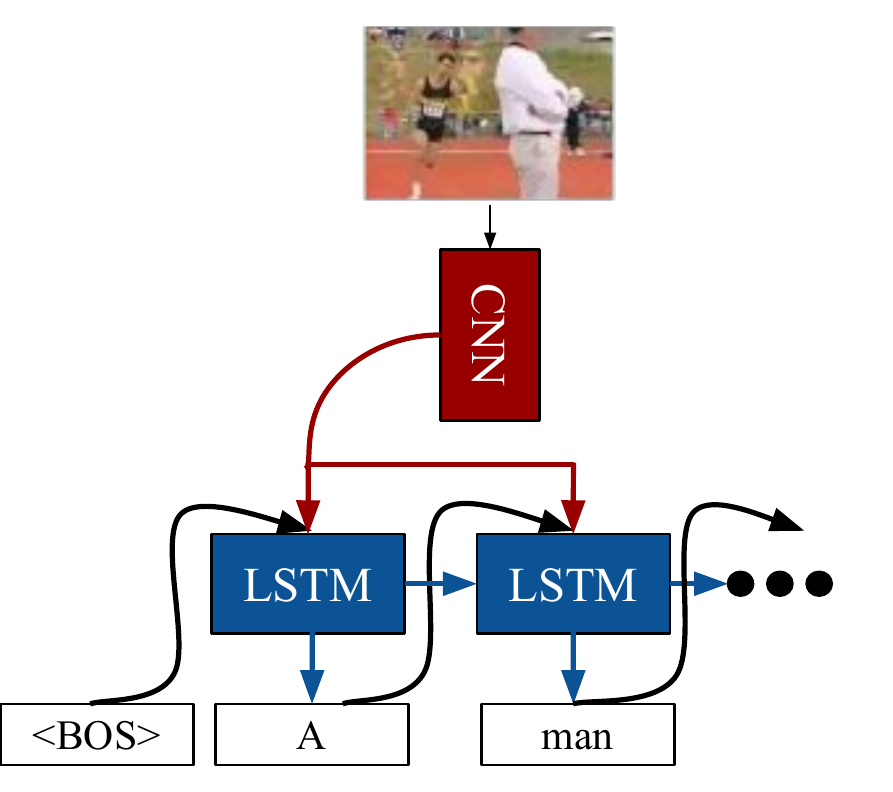}
&
\includegraphics[width=0.25\textwidth]{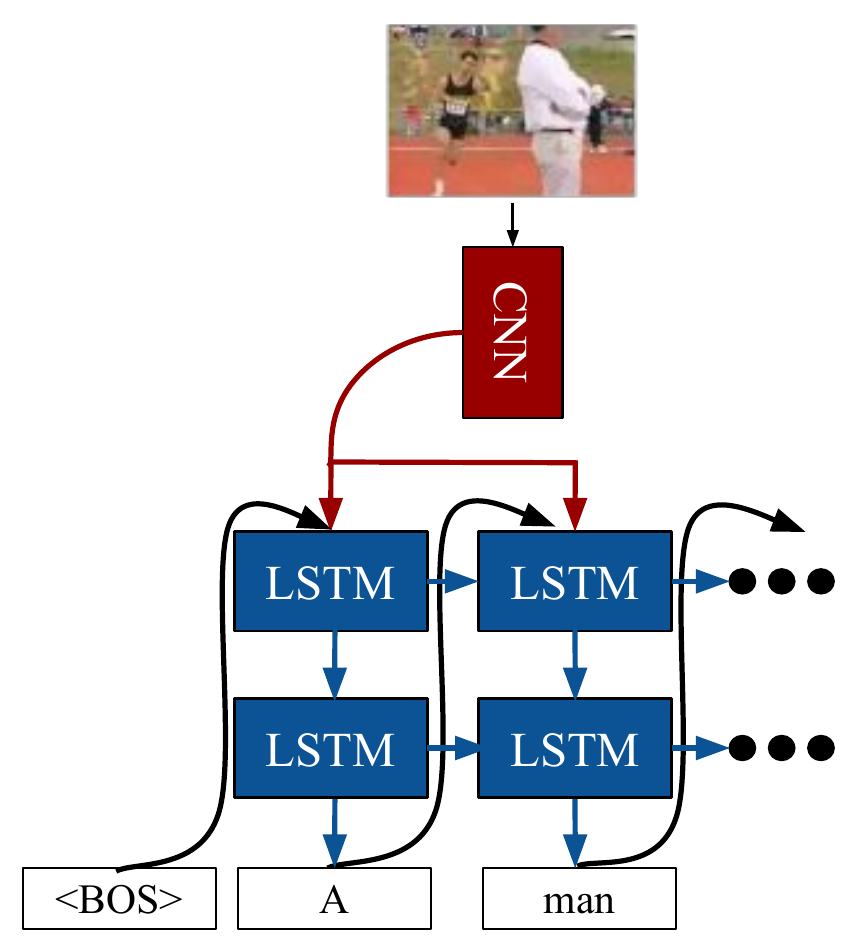}
&
\includegraphics[width=0.25\textwidth]{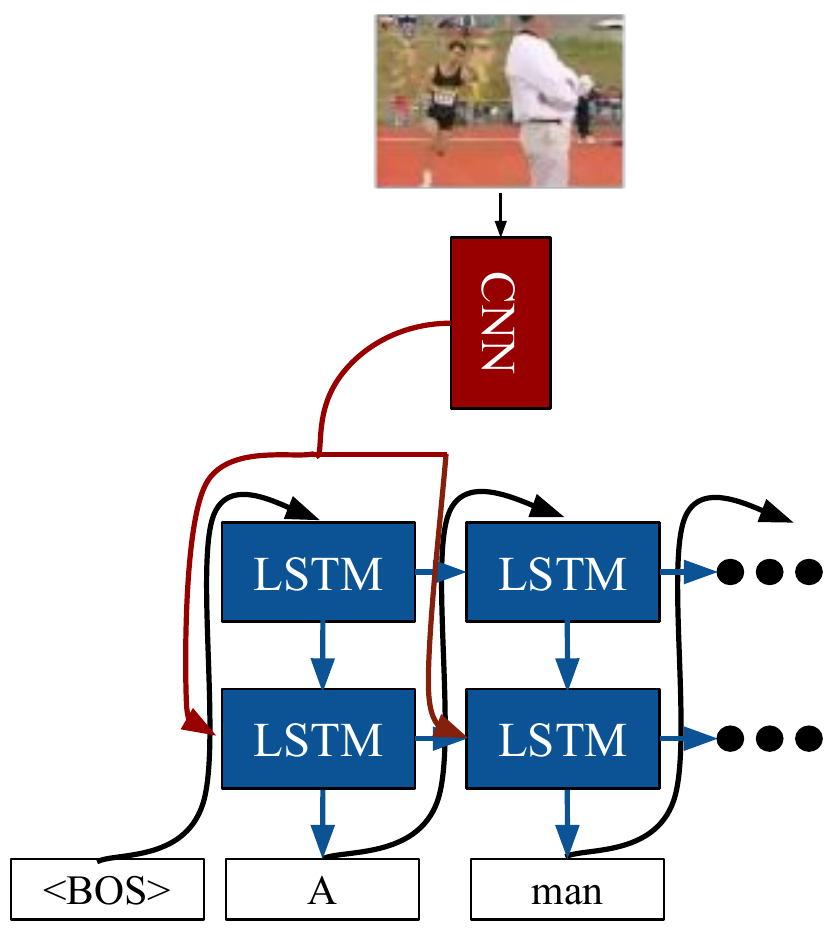}
\\
Single Layer ($L = 1$)
&
Two Layers ($L = 2$),
Unfactored
&
Two Layers ($L = 2$),
Factored
\\
\textbf{LRCN$_{1u}$}
&
\textbf{LRCN$_{2u}$}
&
\textbf{LRCN$_{2f}$}
\\
\end{tabular}
\caption{
Three variants of the LRCN image captioning architecture that we experimentally evaluate.
We explore the effect of depth in the LSTM stack, and the effect of the ``factorization'' of the modalities.
}
\label{fig:captionarchs}
\end{figure*}

The visual feature representation $\phi_V(x)$ of the image $x$ may be input to the sequence model -- a stack of $L$ LSTMs -- by concatenating it at each time step either with
(1) the embedded previous word $W_e y_{t-1}$ and fed into the first LSTM of the stack, or
(2) the hidden state $h_t^{(\ell-1)}$ output from LSTM $\ell-1$ and fed into LSTM $\ell$, for some $\ell \in {2, ..., L}$.
These choices are depicted in Figure~\ref{fig:captionarchs}.
We refer to the latter choice as ``factored,'' as it forces a sort of separation of responsibilities by ``blinding'' the first $\ell-1$ LSTMs and forcing all of the capacity of their hidden states at time step $t$ to represent only the partial caption $y_{1:t-1}$ independent of the visual input, while the LSTMs starting from $\ell$ are responsible for fusing the lower layer's hidden state given by the partial caption with the visual feature representation $\phi_V(x)$ to produce a joint hidden state representation $h_t^{(\ell)}$ of the visual and language inputs up to time step $t$ from which the next word $y_t$ can be predicted.
In the factored case, the hidden state $h_t$ for the lower layers is conditionally independent of the image $x$ given the partial caption $y_{1:t-1}$.

The outputs of the final LSTM in the stack are the inputs to a learned linear prediction layer with a softmax producing a distribution $P(y_t|y_{1:t-1}, \phi_V(x))$ over words $y_t$ in the model's vocabulary, including the \texttt{<EOS>} token denoting the end of the caption, allowing the model to predict captions of varying length.
The visual model $\phi_V$ used for our image captioning experiments is either the \textit{CaffeNet}~\cite{jia2014caffe} reference model, a variant of \textit{AlexNet}~\cite{krizhevsky2012imagenet}, or the more modern and computationally expensive \textit{VGGNet}~\cite{vggnet} model pre-trained for ILSVRC-2012~\cite{ILSVRC15} classification.

Without any explicit language modeling or impositions on the structure of the generated captions, the described LRCN system learns mappings from images input as pixel intensity values to natural language descriptions that are often semantically descriptive and grammatically correct.

At training time, the previous word inputs $y_{1:t-1}$ at time step $t$
are from the \textit{ground truth} caption.
For inference of captions on a novel image $x$, the input is a sample $
\tilde{y}_t
\sim
P(y_t|\tilde{y}_{1:t-1}, \phi_V(x))
$ from the model's predicted distribution at the previous time step,
and generation continues until an \texttt{<EOS>} (end of sequence) token is generated.

\subsection{Evaluation}
We evaluate our image description model for retrieval and generation tasks.
We first demonstrate the effectiveness of our model by quantitatively evaluating it on the
image and caption retrieval tasks proposed by \cite{HodoshEtAl2013} and seen in
\cite{mrnn,karpathy14nips,socher2014grounded,frome2013devise,kirosarXiv}.
We report results on Flickr30k~\cite{flickr30k},
and COCO 2014~\cite{coco2014} datasets, both
with five captions annotated per image.

\subsubsection{Retrieval}
\begin{table}
 \scriptsize
 \begin{center}
 \begin{tabular}{l|cccc}
 \toprule
 & \bf{R@1}  & \bf{R@5} & \bf{R@10} & {\bf{Med}}\textit{r}  \\ \hline
 \midrule
 \multicolumn{5}{c}{Caption to Image (Flickr30k)} \\
 \midrule
 DeViSE \cite{frome2013devise}  & 6.7 & 21.9 & 32.7 & 25 \\
 SDT-RNN \cite{socher2014grounded}  & 8.9 & 29.8 & 41.1 & 16 \\
 DeFrag \cite{karpathy14nips}  & 10.3 & 31.4 & 44.5 & 13 \\
 m-RNN \cite{mrnn}  & 12.6 & 31.2 & 41.5 & 16 \\
 ConvNet \cite{kirosarXiv} & 11.8 & 34.0 & 46.3 & 13\\
 LRCN$_{2f}$ (ours)  & \bf{17.5} & \bf{40.3} & \bf{50.8} & \bf{9} \\
 \midrule
 \multicolumn{5}{c}{Image to Caption (Flickr30k)} \\
 \midrule
 DeViSE \cite{frome2013devise}  & 4.5 & 18.1 & 29.2 & 26 \\
 SDT-RNN \cite{socher2014grounded}  & 9.6 & 29.8 & 41.1 & 16 \\
 DeFrag \cite{karpathy14nips}  & 16.4 & 40.2 & 54.7 & 8 \\
 m-RNN \cite{mrnn}  & 18.4 & 40.2 & 50.9 & 10 \\
 ConvNet \cite{kirosarXiv} & 14.8 & 39.2 & 50.9 & 10 \\
 LRCN$_{2f}$ (ours)  & \bf{23.6} & \bf{46.6} & \bf{58.3} & \bf{7} \\
 \bottomrule
 \end{tabular}
 \end{center}
 \caption{
  Image description: retrieval results for the Flickr30k~\cite{flickr30k} datasets.
 \textbf{R@$K$} is the average recall at rank $K$ (high is good).
 {\bf{Med}}\textit{r} is the median rank (low is good).
}
 \label{tab:imsearchFlickr30k}
 \smallfig
 \end{table}

Retrieval results on the Flickr30k~\cite{flickr30k} dataset
are recorded in Table \ref{tab:imsearchFlickr30k}.
We report median rank, \textbf{Med}\textit{r},
of the first retrieved ground truth image or caption and Recall@$K$,
the number of images or captions for which a
correct caption or image is retrieved within the top $K$ results.
Our model consistently outperforms the strong baselines from recent
work~\cite{kirosarXiv,mrnn,karpathy14nips,socher2014grounded,frome2013devise} as
can be seen in Table \ref{tab:imsearchFlickr30k}.
Here, we note that the \textit{VGGNet} model in \cite{kirosarXiv}
(called \textit{OxfordNet} in their work)
outperforms our model on the retrieval task.
However, \textit{VGGNet} is a stronger
convolutional network~\cite{vggnet} than that used for our results on this task.
The strength of our sequence model (and integration of the sequence and visual models)
can be more directly measured against the \textit{ConvNet}~\cite{kirosarXiv} result,
which uses a very similar base CNN architecture (AlexNet~\cite{krizhevsky2012imagenet}, where we use CaffeNet) pretrained on the same data.

We also ablate the model's retrieval performance on a randomly chosen subset of 1000 images (and 5000 captions) from the COCO 2014~\cite{coco2014} validation set.
Results are recorded in Table~\ref{tab:captionablateretrieval}.
The first group of results for each task examines the effectiveness of an LSTM compared with a ``vanilla'' RNN as described in Section~\ref{sec:background}.
These results demonstrate that the use of the LSTM unit compared to the simpler RNN architecture is an important element of our model's performance on this task,
justifying the additional complexity and suggesting that the LSTM's gating mechanisms allowing for ``long-term'' memory may be quite useful,
even for relatively simple sequences.

Within the second and third result groups, we compare performance among the three sequence model architectural variants depicted in Figure~\ref{fig:captionarchs}.
For both tasks and under all metrics, the two layer, unfactored variant (LRCN$_{2u}$) performs worse than the other two.
The fact that LRCN$_{1u}$ outperforms LRCN$_{2u}$ indicates that stacking additional LSTM layers alone is not beneficial for this task.
The other two variants (LRCN$_{2f}$ and LRCN$_{1u}$) perform similarly across the board, with LRCN$_{2f}$ appearing to have a slight edge in the image to caption task under most metrics,
but the reverse for caption to image retrieval.

Unsurprisingly, finetuning the CNN (indicated by the ``FT?'' column of Table~\ref{tab:captionablateretrieval}) and using a more powerful CNN (VGGNet~\cite{vggnet} rather than CaffeNet) each improve results substantially across the board.
Finetuning boosts the R@$k$ metrics by 3-5\% for CaffeNet, and 5-8\% for VGGNet.
Switching from CaffeNet to VGGNet improves results by around 8-12\% for the caption to image task, and by roughly 11-17\% for the image to caption task.

\begin{table}
\scriptsize
\begin{center}
\begin{tabular}{cc|ccc|cccc}
\toprule
\multicolumn{2}{c}{Vision Model}
&
\multicolumn{3}{c}{Sequence Model}
&
\multicolumn{4}{c}{Retrieval Performance}
\\
\midrule
CNN & FT? & Unit & $L$ & Factor? &
R@1  & R@5 & R@10 & Med$r$  \\ \hline
\midrule
&&&&&
\multicolumn{4}{c}{Caption to Image} \\
\midrule
CaffeNet & - & RNN & 2 & \checkmark &
21.3 & 51.7 & 67.2 & 5 \\
CaffeNet & - & LSTM & 2 & \checkmark &
\bf{25.0} & \bf{56.2} & \bf{70.6} & \bf{4} \\
\midrule
CaffeNet & - & LSTM & 1 & - &
\bf{25.2} & \bf{56.2} & \bf{70.8} & \bf{4} \\
CaffeNet & - & LSTM & 2 & - &
23.4 & 54.8 & 69.3 & 5 \\
CaffeNet & - & LSTM & 2 & \checkmark &
25.0 & \bf{56.2} & 70.6 & \bf{4} \\
\midrule
CaffeNet & \checkmark & LSTM & 1 & - &
\bf{28.5} & \bf{60.0} & 74.5 & \bf{4} \\
CaffeNet & \checkmark & LSTM & 2 & - &
25.6 & 57.2 & 72.2 & \bf{4} \\
CaffeNet & \checkmark & LSTM & 2 & \checkmark &
27.2 & 59.6 & \bf{74.7} & \bf{4} \\
\midrule
VGGNet & - & LSTM & 2 & \checkmark &
33.5 & 68.1 & 80.8 & 3 \\
VGGNet & \checkmark & LSTM & 2 & \checkmark &
\bf{39.3} & \bf{74.7} & \bf{85.9} & \bf{2} \\
\midrule
&&&&&
\multicolumn{4}{c}{Image to Caption} \\
\midrule
CaffeNet & - & RNN & 2 & \checkmark &
30.2 & 61.0 & 72.6 & 4 \\
CaffeNet & - & LSTM & 2 & \checkmark &
\bf{33.8} & \bf{65.3} & \bf{75.3} & \bf{3} \\
\midrule
CaffeNet & - & LSTM & 1 & - &
32.3 & 64.5 & \bf{75.6} & \bf{3} \\
CaffeNet & - & LSTM & 2 & - &
29.9 & 60.8 & 72.7 & \bf{3} \\
CaffeNet & - & LSTM & 2 & \checkmark &
\bf{33.8} & \bf{65.3} & 75.3 & \bf{3} \\
\midrule
CaffeNet & \checkmark & LSTM & 1 & - &
36.1 & \bf{68.4} & 79.5 & 3 \\
CaffeNet & \checkmark & LSTM & 2 & - &
33.1 & 63.7 & 76.9 & 3 \\
CaffeNet & \checkmark & LSTM & 2 & \checkmark &
\bf{36.3} & 67.3 & \bf{80.6} & \bf{2} \\
\midrule
VGGNet & - & LSTM & 2 & \checkmark &
46.0 & 77.4 & 88.3 & 2 \\
VGGNet & \checkmark & LSTM & 2 & \checkmark &
\bf{53.3} & \bf{84.3} & \bf{91.9} & \bf{1} \\
\bottomrule
\end{tabular}
\end{center}
\caption{ Retrieval results (image to caption and caption to image) for a randomly chosen subset (1000 images) of the COCO 2014~\cite{coco2014} validation set.
R@$K$ is the average recall at rank $K$ (high is good).
Med\textit{r} is the median rank (low is good).
}
\label{tab:captionablateretrieval}
\smallfig
\end{table}

\begin{table*}
\scriptsize
\begin{center}
\begin{tabular}{ccc|cc|ccc|ccccccc}
\toprule
\multicolumn{3}{c}{Generation Strategy}
&
\multicolumn{2}{c}{Vision Model}
&
\multicolumn{3}{c}{Sequence Model}
&
\multicolumn{7}{c}{Generation Performance (COCO 2014~\cite{coco2014} Validation Set)}
\\
\midrule
Beam & \multicolumn{2}{c|}{Sample}
& & & & &
\\
Width & $N$ & $T$ &
CNN & FT? & Unit & $L$ & Factor? &
B1 & B2 & B3 & B4 & C & M & R
\\
\hline
\midrule
1 & - & - & CaffeNet & - & RNN & 2 & \checkmark &
0.638&0.454&0.315&0.220&0.660&0.209&0.473
\\
1 & - & - & CaffeNet & - & LSTM & 2 & \checkmark &
\bf{0.646}&\bf{0.462}&\bf{0.321}&\bf{0.224}&\bf{0.674}&\bf{0.210}&\bf{0.477}
\\
\midrule
1 & - & - & CaffeNet & - & LSTM & 1 & - &
\bf{0.654}&\bf{0.475}&\bf{0.333}&\bf{0.231}&0.661&0.209&\bf{0.480}
\\
1 & - & - & CaffeNet & - & LSTM & 2 & - &
0.653&0.470&0.328&0.230&\bf{0.682}&\bf{0.212}&\bf{0.480}
\\
1 & - & - & CaffeNet & - & LSTM & 2 & \checkmark &
0.646&0.462&0.321&0.224&0.674&0.210&0.477
\\
\midrule
1 & - & - & CaffeNet & \checkmark & LSTM & 1 & - &
\bf{0.661}&\bf{0.485}&\bf{0.344}&\bf{0.241}&0.702&0.216&\bf{0.489}
\\
1 & - & - & CaffeNet & \checkmark & LSTM & 2 & - &
0.659&0.478&0.338&0.238&0.716&0.217&0.486
\\
1 & - & - & CaffeNet & \checkmark & LSTM & 2 & \checkmark &
0.659&0.478&0.336&0.237&\bf{0.717}&\bf{0.218}&0.486
\\
\midrule
1 & - & - & VGGNet & - & LSTM & 2 & \checkmark &
0.674&0.494&0.351&0.248&0.773&0.227&0.497
\\
1 & - & - & VGGNet & \checkmark & LSTM & 2 & \checkmark &
\bf{0.695}&\bf{0.519}&\bf{0.374}&\bf{0.268}&\bf{0.839}&\bf{0.237}&\bf{0.512}
\\
\hline
\midrule
- & 100 & 1.5 & CaffeNet & - & RNN & 2 & \checkmark &
0.647&0.466&0.334&0.244&0.703&0.212&0.479
\\
- & 100 & 1.5 & CaffeNet & - & LSTM & 2 & \checkmark &
\bf{0.657}&\bf{0.478}&\bf{0.344}&\bf{0.251}&\bf{0.720}&\bf{0.215}&\bf{0.485}
\\
\midrule
- & 100 & 1.5 & CaffeNet & - & LSTM & 1 & - &
\bf{0.664}&\bf{0.490}&\bf{0.354}&0.254&0.704&0.211&0.488
\\
- & 100 & 1.5 & CaffeNet & - & LSTM & 2 & - &
\bf{0.664}&0.486&0.352&\bf{0.257}&\bf{0.732}&\bf{0.216}&\bf{0.489}
\\
- & 100 & 1.5 & CaffeNet & - & LSTM & 2 & \checkmark &
0.657&0.478&0.344&0.251&0.720&0.215&0.485
\\
\midrule
- & 100 & 1.5 & CaffeNet & \checkmark & LSTM & 1 & - &
\bf{0.679}&\bf{0.507}&\bf{0.370}&\bf{0.268}&0.753&0.219&\bf{0.499}
\\
- & 100 & 1.5 & CaffeNet & \checkmark & LSTM & 2 & - &
0.672&0.495&0.361&0.265&0.762&\bf{0.222}&0.495
\\
- & 100 & 1.5 & CaffeNet & \checkmark & LSTM & 2 & \checkmark &
0.670&0.493&0.358&0.264&\bf{0.764}&\bf{0.222}&0.495
\\
\midrule
- & 100 & 1.5 & VGGNet & - & LSTM & 2 & \checkmark &
0.690&0.514&0.377&0.278&0.828&0.231&0.508
\\
- & 100 & 1.5 & VGGNet & \checkmark & LSTM & 2 & \checkmark &
\bf{0.711}&\bf{0.541}&\bf{0.402}&\bf{0.300}&\bf{0.896}&\bf{0.242}&\bf{0.524}
\\
\hline
\midrule
1 & - & - &
VGGNet & \checkmark & LSTM & 2 & \checkmark &
0.695&0.519&0.374&0.268&0.839&0.237&0.512
\\
2 & - & - &
VGGNet & \checkmark & LSTM & 2 & \checkmark &
0.707&0.533&0.394&0.291&0.879&0.242&0.520
\\
3 & - & - &
VGGNet & \checkmark & LSTM & 2 & \checkmark &
\bf{0.708}&\bf{0.536}&\bf{0.399}&0.298&\bf{0.888}&\bf{0.243}&\bf{0.521}
\\
4 & - & - &
VGGNet & \checkmark & LSTM & 2 & \checkmark &
0.706&0.534&0.398&0.299&\bf{0.888}&\bf{0.243}&\bf{0.521}
\\
5 & - & - &
VGGNet & \checkmark & LSTM & 2 & \checkmark &
0.704&0.533&0.398&\bf{0.300}&\bf{0.888}&0.242&0.520
\\
10 & - & - &
VGGNet & \checkmark & LSTM & 2 & \checkmark &
0.699&0.528&0.395&0.298&0.886&0.241&0.518
\\
\midrule
- & 1 & 2.0 &
VGGNet & \checkmark & LSTM & 2 & \checkmark &
0.658&0.472&0.327&0.224&0.733&0.222&0.483
\\
- & 10 & 2.0 &
VGGNet & \checkmark & LSTM & 2 & \checkmark &
0.708&0.534&0.391&0.286&0.868&0.239&0.519
\\
- & 25 & 2.0 &
VGGNet & \checkmark & LSTM & 2 & \checkmark &
0.712&0.540&0.398&0.294&0.885&0.241&0.523
\\
- & 100 & 2.0 &
VGGNet & \checkmark & LSTM & 2 & \checkmark &
\bf{0.714}&\bf{0.543}&\bf{0.402}&\bf{0.297}&\bf{0.889}&\bf{0.242}&\bf{0.524}
\\
\midrule
- & 100 & 1.0 &
VGGNet & \checkmark & LSTM & 2 & \checkmark &
0.674&0.494&0.357&0.261&0.805&0.228&0.494
\\
- & 100 & 1.5 &
VGGNet & \checkmark & LSTM & 2 & \checkmark &
0.711&0.541&\bf{0.402}&\bf{0.300}&\bf{0.896}&\bf{0.242}&\bf{0.524}
\\
- & 100 & 2.0 &
VGGNet & \checkmark & LSTM & 2 & \checkmark &
\bf{0.714}&\bf{0.543}&\bf{0.402}&0.297&0.889&\bf{0.242}&\bf{0.524}
\\
\bottomrule
\end{tabular}
\end{center}
\caption{ Image caption generation performance (under the BLEU
1-4~\cite{papineni2002bleu} (B1-B4), CIDEr-D~\cite{cider} (C),
METEOR~\cite{banerjee2005meteor} (M), and ROUGE-L~\cite{lin2004rouge} (R) metrics) across various network architectures and generation strategies.
In the topmost set of results, we show performance across various CNN and recurrent architectures for a simple generation strategy -- beam search with beam width 1 (\ie, simply choosing the most probable word at each time step).
In the middle set of results, we show performance across the same set of architectures for a more sophisticated and computationally intensive generation strategy found to be the best performing (in terms of performance under the CIDEr-D metric) among those explored in the bottom-most set of results, which explores various generation strategies while fixing the choice of network.
In the first two sets of results, we vary the visual input CNN architecture
(either CaffeNet~\cite{jia2014caffe}, an architecture similar to
AlexNet~\cite{krizhevsky2012imagenet}, or the more modern VGGNet~\cite{vggnet}) and whether its weights are finetuned (FT?).
Keeping the visual input CNN fixed with CaffeNet, we also vary the choice of recurrent architecture, comparing a stack of ``vanilla'' RNNs with LSTMs~\cite{hochreiter1997long}, as well as the number of layers in the stack $L$, and (for $L=2$) whether the layers are ``factored'' (\ie, whether the visual input is passed into the second layer).
In the last set of results, we explore two generation strategies -- beam search, and choosing the best (highest log-likelihood) among $N$ samples from the model's predicted distribution.
For beam search we vary the beam width from 1-10.
For the sampling strategy we explore the effect of sample size $N$ as well as the effect of applying various choices of scalar factor $T$ (inverse of the ``temperature'') to the logits input to the softmax producing the distribution.
}
\label{tab:captiongenablate}
\end{table*}

\begin{table*}
\scriptsize
\begin{center}
\begin{tabular}{ccl|ccccccc}
\toprule
& & &
\multicolumn{7}{c}{Generation Performance (COCO 2014~\cite{coco2014} Test Set)}
\\
\midrule
\multicolumn{3}{c|}{Method} &
B1 & B2 & B3 & B4 & C & M & R
\\
\hline
\midrule
& \cite{vinyals2014show} & NIC			 &     0.895  &     0.802  &     0.694  &     0.587  & \bf{0.946} & \bf{0.346} & \bf{0.682} \\
& \cite{captivator} & MSR Captivator	 & \bf{0.907} & \bf{0.819} & \bf{0.710} & \bf{0.601} &     0.937  &     0.339  &     0.680  \\
& \cite{mao2015learning} & m-RNN (2015)	 &     0.890  &     0.798  &     0.687  &     0.575  &     0.935  &     0.325  &     0.666  \\
(Ours) & * & LRCN, this work (sample)
					 &     0.895  &     0.804  &     0.695  &     0.585  &     0.934  &     0.335  &     0.678  \\
& \cite{msr} & MSR			 &     0.880  &     0.789  &     0.678  &     0.567  &     0.925  &     0.331  &     0.662  \\
& \cite{nearestneigh} & Nearest Neighbor &     0.872  &     0.770  &     0.655  &     0.542  &     0.916  &     0.318  &     0.648  \\
& \cite{coco2014} & Human		 &     0.880  &     0.744  &     0.603  &     0.471  &     0.910  &     0.335  &     0.626  \\
& \cite{mrnn} & m-RNN (2014)		 &     0.890  &     0.801  &     0.690  &     0.578  &     0.896  &     0.320  &     0.668  \\
(Ours) & \cite{lrcn} & LRCN (greedy)
					 &     0.871  &     0.772  &     0.653  &     0.534  &     0.891  &     0.322  &     0.656  \\
& \cite{sat} & Show, Attend, and Tell	 &     0.872  &     0.768  &     0.644  &     0.523  &     0.878  &     0.323  &     0.651  \\
& \cite{kirosarXiv} & MLBL		 &     0.848  &     0.747  &     0.633  &     0.517  &     0.752  &     0.294  &     0.635  \\
& \cite{karpathy2014deep} & NeuralTalk	 &     0.828  &     0.701  &     0.566  &     0.446  &     0.692  &     0.280  &     0.603  \\
\bottomrule
\end{tabular}
\end{center}
\caption{
Image caption generation results from top-performing methods in the 2015 COCO caption challenge competition, sorted by performance under the CIDEr-D metric.
(We omit submissions that did not provide a reference to a report describing their method; see full results at \protect\url{http://mscoco.org/dataset/\#captions-leaderboard}.)
All results except for our updated result (denoted by \textit{LRCN, this work}) were competition entries (submitted by May 2015).
Our updated result differs from our original competition entry only by generation strategy (sampling with $N=100$, $T=1.5$, rather than beam search with width 1; \ie, greedy search); the visual and recurrent architectures (and trained weights) are the same.
}
\label{tab:captiongentest}
\end{table*}

\subsubsection{Generation}
We evaluate LRCN's caption generation performance on the COCO2014~\cite{coco2014} dataset
using the official metrics on which COCO image captioning submissions are evaluated.
The BLEU \cite{papineni2002bleu} and METEOR~\cite{banerjee2005meteor} metrics were designed for automatic evaluation of
machine translation methods.
ROUGE-L~\cite{lin2004rouge} was designed for evaluating summarization performance.
CIDEr-D~\cite{cider} was designed specifically to evaluate the image captioning task.

In Table~\ref{tab:captiongenablate} we evaluate variants of our model along the same axes as done for the retrieval tasks in Table~\ref{tab:captionablateretrieval}.
In the last of the three groups of results, we additionally explore and evaluate various caption generation strategies that can be employed for a given network.
The simplest strategy, and the one employed for most of our generation results in our prior work~\cite{lrcn}, is to generate captions greedily; \ie, by simply choosing the most probable word at each time step.
This is equivalent to (and denoted in Table~\ref{tab:captiongenablate} by) beam search with beam width 1.
In general, beam search with beam width $N$ approximates the most likely caption by retaining and expanding only the $N$ current most likely partial captions, according to the model.
We find that of the beam search strategies, a beam width of 3-5 gives the best generation numbers -- performance saturates quickly and even degrades for larger beam width (e.g., 10).

An alternative, non-deterministic generation strategy is to randomly sample $N$ captions from the model's distribution and choose the most probable among these.
Under this strategy we also examine the effect of applying various choices of scalar factors (inverse of the ``temperature'') $T$ to the real-valued predictions input to the softmax producing the distribution.
For larger values of $T$ the samples are greedier and less diverse, with $T = \infty$ being equivalent to beam search with beam width 1.
Larger values of $N$ suggest using smaller values of $T$, and vice versa -- for example, with large $N$ and large $T$, most of the $\mathcal{O}(N)$ computation is wasted as many of the samples will be redundant.
We assess saturation as the number of samples $N$ grows, and find that $N=100$ samples with $T=2$ improves little over $N=25$.
We also varied the temperature $T$ among values 1, 1.5, and 2 (all with $N=100$) and found $T=1.5$ to perform the best.

We adopt the best-performing generation strategy from the bottom-most set of results in Table~\ref{tab:captiongenablate} (sampling with $T=1.5$, $N=100$) as the strategy for the middle set of results in the table, which ablates LRCN architectures.
We also record generation performance for all architectures (Table~\ref{tab:captiongenablate}, top set of results) with the simpler generation strategy used in our earlier work~\cite{lrcn} for ease of comparison with this work and for future researchers.
For the remainder of this discussion, we will focus on the middle set of results, and particularly on the CIDEr-D~\cite{cider} (C) metric, as it was designed specifically for automatic evaluation of image captioning systems.
We see again that the LSTM unit outperforms an RNN unit for generation, though not as significantly as for retrieval.
Between the sequence model architecture choices (depicted in Figure~\ref{fig:captionarchs}) of the number of layers $L$ and whether to factor,
we see that in this case the two-layer models (LRCN$_{2f}$ and LRCN$_{2u}$) perform similarly, outperforming the single layer model (LRCN$_{1u}$).
Interestingly, of the three variants, LRCN$_{2f}$ is the only one to perform best for both retrieval and generation.

We see again that fine-tuning (FT) the visual representation and using a stronger vision model (VGGNet~\cite{vggnet}) improves results significantly.
Fine-tuning improves CIDEr-D by roughly 0.04 points for CaffeNet, and by roughly 0.07 points for VGGNet.
Switching from finetuned CaffeNet to VGGNet improves CIDEr-D by 0.13 points.

In Table~\ref{tab:captiongentest} we compare generation performance with contemporaneous and recent work submitted to the 2015 COCO caption challenge using our best-performing method (under the CIDEr-D metric) from the results on the validation set described above -- generating a caption for a single image by taking the best of $N=100$ samples with a scalar factor of $T=1.5$ applied to the softmax inputs, using an LRCN model which pairs a fine-tuned VGGNet with our LRCN$_{2f}$ (two layer, factored) sequence model architecture.
Our results are competitive with the contemporary work, performing 4th best in CIDEr-D (0.934, compared with the best result of 0.946 from~\cite{vinyals2014show}), and 3rd best in METEOR (0.335, compared with 0.346 from~\cite{vinyals2014show}).

In addition to standard quantitative
evaluations, we also employ Amazon Mechnical Turk workers (``Turkers'') to evaluate the
generated sentences. Given an image and a set of descriptions from different
models, we ask Turkers to rank the sentences based on correctness, grammar and
relevance. We compared sentences from our model to the ones made publicly available by \cite{kirosarXiv}. 
As seen in Table~\ref{tab:humaneval}, our fine-tuned (FT) LRCN model
performs on par with the Nearest Neighbour (NN) on correctness and relevance, and better on grammar.

\begin{figure}
\begin{center}
\begin{tabular}{p{5.5cm}p{5.5cm}p{5.5cm}}
\includegraphics[width=5.5cm,height=4cm]{./coco_images/COCO_val2014_000000411438.jpg}&
\includegraphics[width=5.5cm,height=4cm]{./coco_images/COCO_val2014_000000438862.jpg}&
\includegraphics[width=5.5cm,height=4cm]{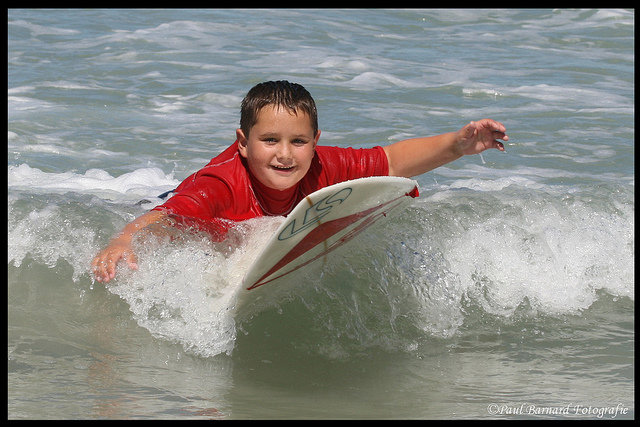}\\
A female tennis player in action on the court.&
A group of young men playing a game of soccer& 
A man riding a wave on top of a surfboard.\\\\
\includegraphics[width=5.5cm,height=4cm]{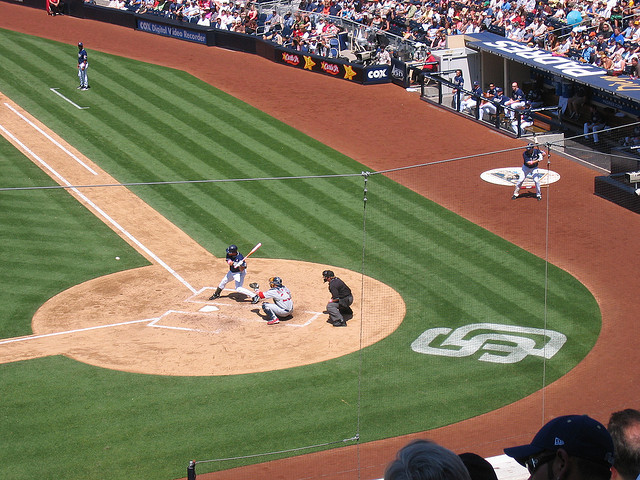} &
\includegraphics[width=5.5cm,height=4cm]{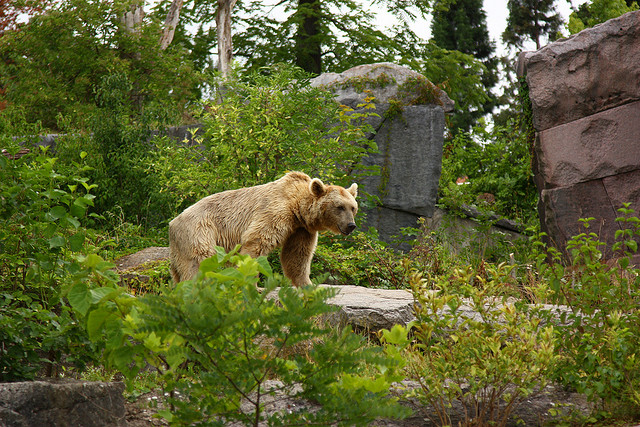} &
\includegraphics[width=5.5cm,height=4cm]{./coco_images/COCO_val2014_000000499266.jpg}\\
A baseball game in progress with the batter up to plate.&
A brown bear standing on top of a lush green field.&
A person holding a cell phone in their hand.\\\\
\includegraphics[width=5.5cm,height=4cm]{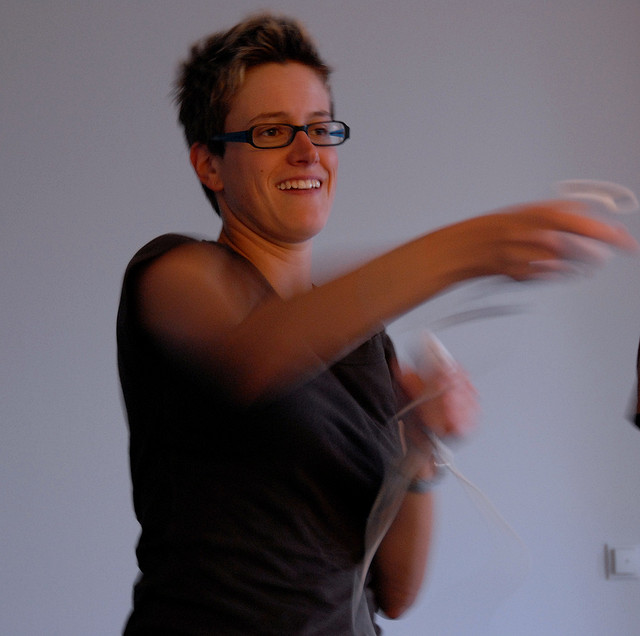}&
\includegraphics[width=5.5cm,height=4cm]{./coco_images/COCO_val2014_000000438861.jpg}&
\includegraphics[width=5.5cm,height=4cm]{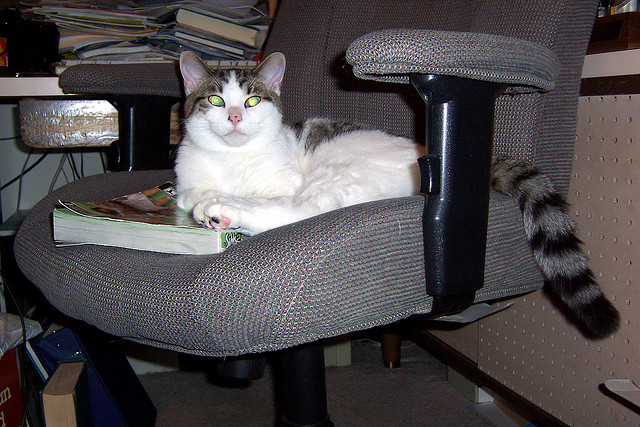}\\
 A close up of a person brushing his teeth.&
A woman laying on a bed in a bedroom.&
A black and white cat is sitting on a chair.\\\\
\includegraphics[width=5.5cm,height=4cm]{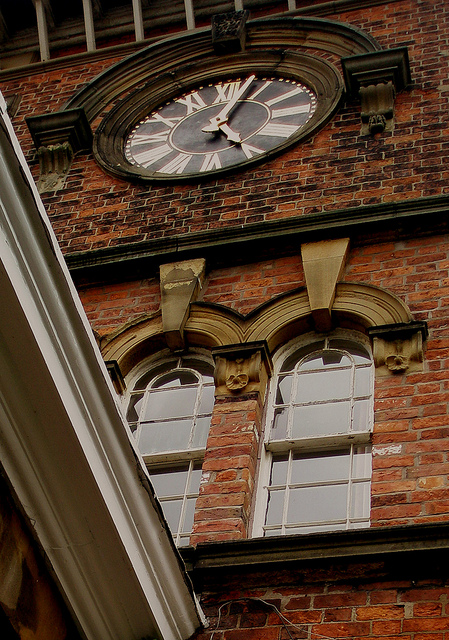}&
\includegraphics[width=5.5cm,height=4cm]{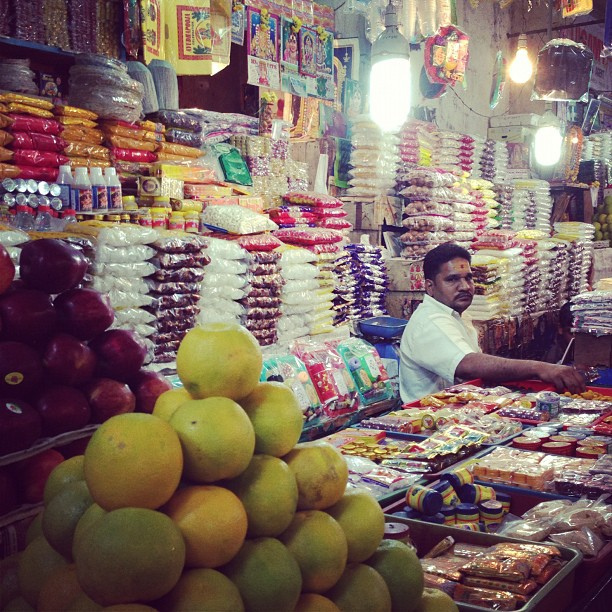}&
\includegraphics[width=5.5cm,height=4cm]{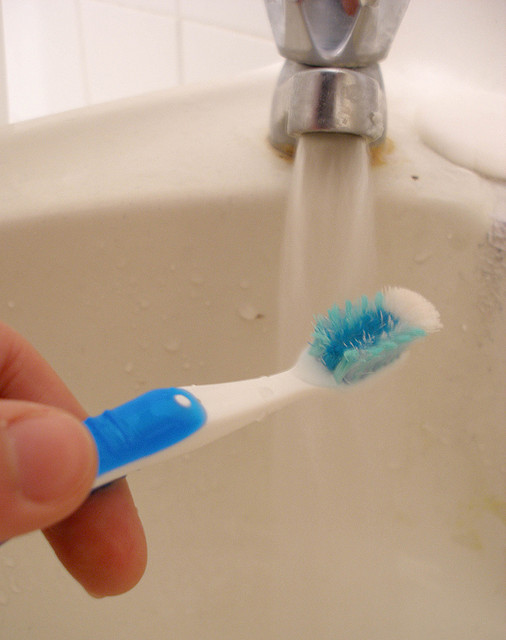}\\
A large clock mounted to the side of a building.&
A bunch of fruit that are sitting on a table.&
A toothbrush holder sitting on top of a white sink.\\
\end{tabular}
\end{center}
\caption{Image description: images with corresponding captions generated by our finetuned LRCN model.
These are images 1-12 of our randomly chosen validation set from COCO 2014~\cite{coco2014}.
We used beam search with a beam size of 5 to generate the sentences, and display the top (highest likelihood) result above.
}
\label{fig:exampleAnnotations}

\end{figure}

We show sample captions in Figure~\ref{fig:exampleAnnotations}.
We additionally note some properties of the captions our model generates.
When using the VGG model to generate sentences in the validation set, we find that 33.7\% of our generated setences exactly match a sentence in the training set.
Furthermore, we find that when using a beam size of one, our model generates 42\% of the vocabulary words used by human annotators when describing images in the validation set.
Some words, such as ``lady'' and ``guy'', are not generated by our model but are commonly used by human annotators, but synonyms such as ``woman'' and ``man'' are two of the most common words generated by our model.

\begin{table}
\scriptsize
\begin{center}
\begin{tabular}{lccc}
\toprule
& Correctness & Grammar & Relevance\\
\midrule
TreeTalk \cite{kuznetsova2014treetalk}	&4.08	&4.35	&3.98	\\
VGGNet \cite{kirosarXiv}	&3.71	&3.46	&3.70	\\
NN \cite{kirosarXiv}	&\bf{3.44}	&3.20	&\bf{3.49}	\\
LRCN fc$_8$ (ours)	&3.74	&3.19	&3.72	\\
LRCN FT (ours)	&\bf{3.47}	&\bf{3.01}	&\bf{3.50}	\\
\midrule
Captions	&2.55	&3.72	&2.59	\\
\bottomrule
\end{tabular}
\end{center}
\caption{Image description: 
Human evaluator rankings from 1-6 (low is good) averaged for each method and criterion.
We evaluated on 785 Flickr images selected by the authors of~\cite{kirosarXiv} for the purposes of comparison against this similar contemporary approach.
}
\label{tab:humaneval}
\smallfig
\end{table}

\section{Video description}
\label{sec:videoDescription}
\begin{figure*}
\centering
\begin{tabular}{c|c|c}
\includegraphics[width=0.31\textwidth]{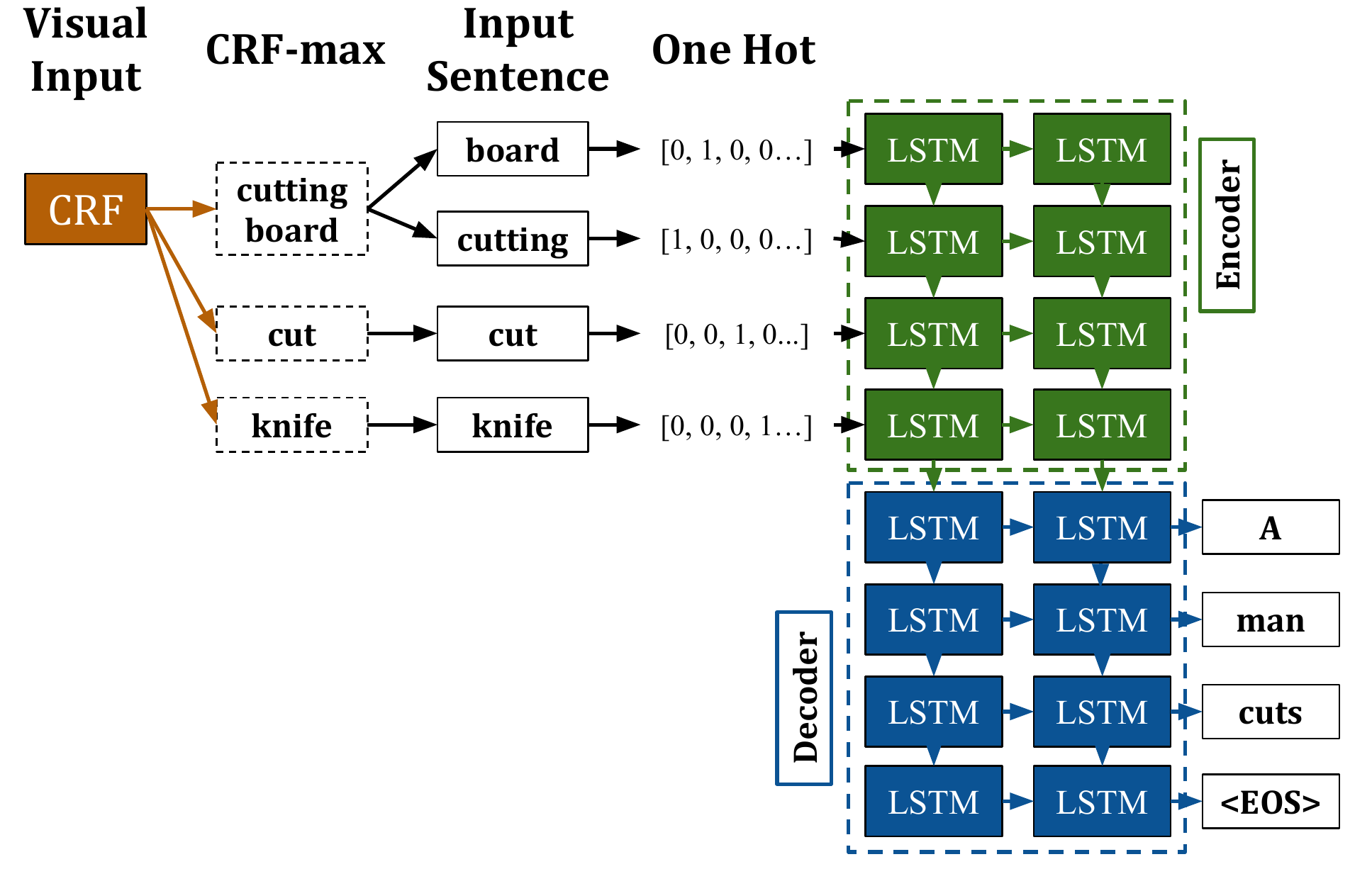}
&
\includegraphics[width=0.31\textwidth]{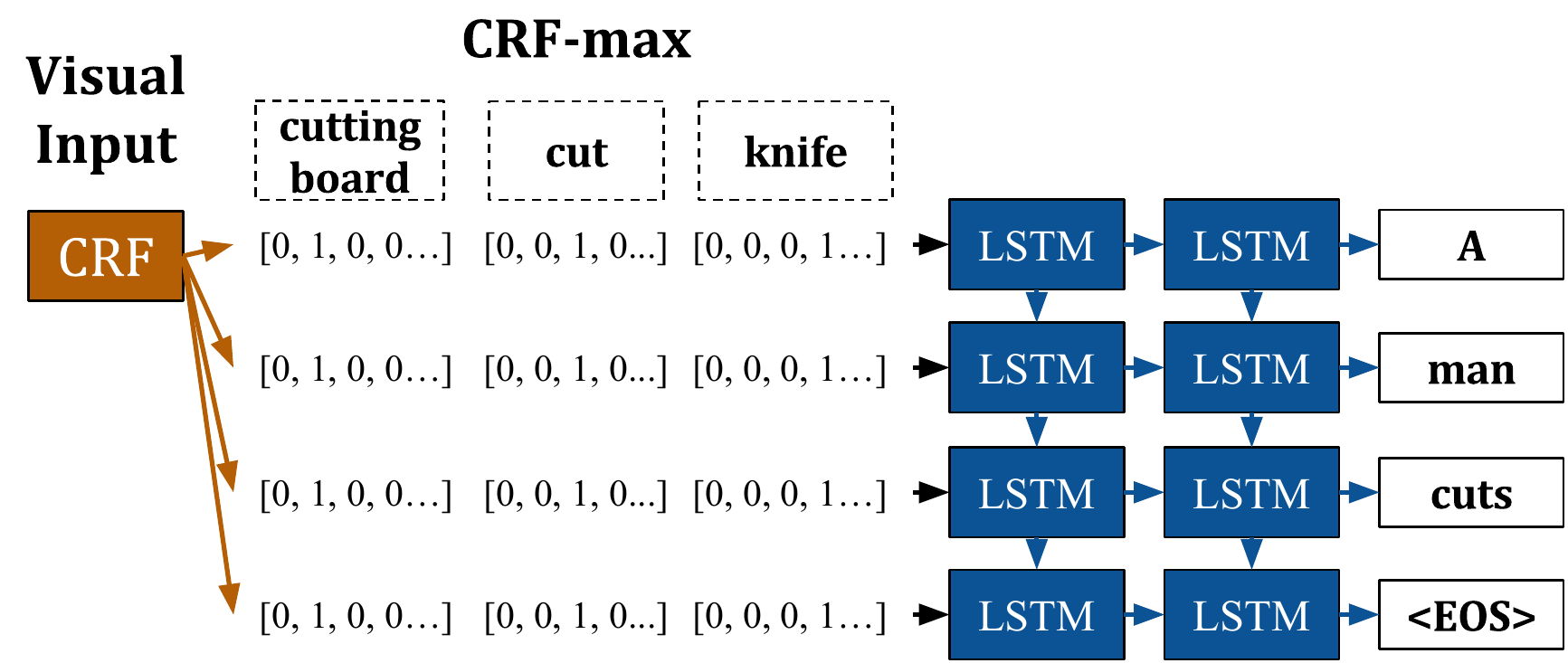}
&
\includegraphics[width=0.31\textwidth]{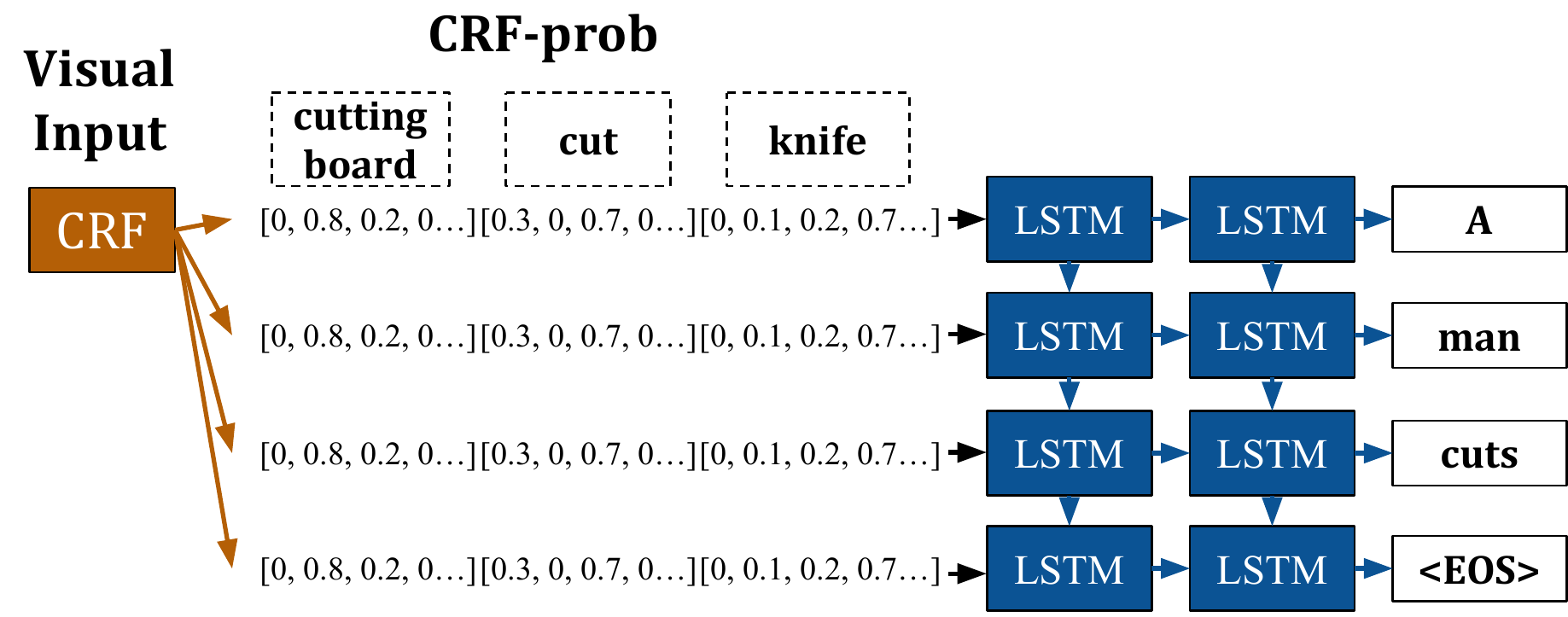}
\\
(a)
&
(b)
&
(c)
\\
\textbf{LSTM Encoder-Decoder}
&
\textbf{LSTM Decoder (CRF-max)}
&
\textbf{LSTM Decoder (CRF-prob)}
\\
\end{tabular}
  \caption{Our approaches to video description.   (a) LSTM encoder \& decoder with CRF max (b) LSTM decoder with CRF max (c) LSTM decoder with CRF probabilities.}
  \label{fig:videoDescription:appraoch}
\vspace{-0.1in}
\end{figure*}

In video description
the LSTM framework allows us to model the video as a variable length input stream. 
However, due to the limitations of available video description datasets, we rely on more ``traditional'' activity and video recognition processing for the input and use LSTMs for generating a sentence.
We first distinguish the following architectures for video description (see Figure~\ref{fig:videoDescription:appraoch}).
For each architecture, we assume we have predictions of activity, tool, object, and locations present in the video from a CRF based on the full video input. In this way, we observe the video as whole at each time step, not incrementally frame by frame.
\label{sec:videoDescription:approach}

\textbf{(a) LSTM encoder \& decoder with CRF max.} (Figure~\ref{fig:videoDescription:appraoch}(a))
This architecture is motivated by the video description approach presented in \cite{rohrbach13iccv}. 
They first recognize a semantic representation of the video using the maximum a posteriori (MAP) estimate of a CRF with video features as unaries.
This representation, \eg, $\langle$knife,cut,carrot,cutting board$\rangle$, is concatenated into an input sequence (\emph{knife cut carrot cutting board}) which is translated to a natural language sentence (\emph{a person cuts a carrot on the board}) using statistical machine translation (SMT)  \cite{koehn07acl}. 
We replace SMT with an encoder-decoder LSTM, which encodes the input sequence as a fixed length vector before decoding to a sentence.

\textbf{(b) LSTM decoder with CRF max.} (Figure~\ref{fig:videoDescription:appraoch}(b))
In this variant we provide the full visual input representation at each time step to the LSTM, analogous to how an image is provided as an input to the LSTM in image captioning.  

\textbf{(c) LSTM decoder with CRF probabilites.} (Figure~\ref{fig:videoDescription:appraoch}(c))
A benefit of using LSTMs for machine translation compared to phrase-based SMT \cite{koehn07acl} is that it can naturally incorporate probability vectors during training \emph{and} test time which allows the LSTM to learn uncertainties in visual generation rather than relying on MAP estimates. The architecture is the the same as in (b), but we replace max predictions  with probability distributions.

\begin{table}
\centering
\scriptsize
\begin{tabular}{ll@{}r}
\toprule
Architecture & Input & BLEU\\
\midrule
SMT \cite{rohrbach13iccv} &  CRF max & 24.9 \\
SMT \cite{rohrbach2014gcpr} &  CRF prob & 26.9 \\
(a) LSTM Encoder-Decoder (ours) & CRF max & 25.3\\
(b) LSTM Decoder (ours) & CRF max  & 27.4\\
(c) LSTM Decoder (ours) & CRF prob & \bf{28.8} \\
\bottomrule
\end{tabular}
\vspace{1.0mm}
\caption{Video description: Results on detailed description of TACoS multilevel\cite{rohrbach2014gcpr}, in \%, see Section \ref{sec:videoDescription:approach} for details.}
\label{tbl:videodesc:res}
\smallfig
\end{table}

\subsection{Evaluation}
We evaluate our approach on the TACoS multilevel \cite{rohrbach2014gcpr} dataset, which has 44,762 video/sentence pairs (about 40,000 for training/validation). We compare to \cite{rohrbach13iccv} who use max prediction as well as a variant presented in \cite{rohrbach2014gcpr} which takes CRF probabilities at test time and uses a word lattice to find an optimal sentence prediction.
Since we use the max prediction as well as the probability scores provided by \cite{rohrbach2014gcpr}, we have an identical visual representation. \cite{rohrbach2014gcpr} uses dense trajectories \cite{wang13ijcv} and SIFT features as well as temporal context reasoning modeled in a CRF.
In this set of experiments we use the two-layered, unfactored version of LRCN, as described for image description.

Table~\ref{tbl:videodesc:res} shows the BLEU-4 score. The results show that (1) the LSTM outperforms an SMT-based approach to video description; (2) the simpler decoder architecture (b) and (c) achieve better performance than (a), likely because the input does not need to be memorized; and (3) our approach achieves 28.8\%, clearly outperforming the best reported number of 26.9\% on TACoS multilevel by \cite{rohrbach2014gcpr}.

More broadly, these results show that our architecture is not restricted only to input from deep networks,
but can be cleanly integrated with fixed or variable length inputs from other vision systems.

\section{Related Work}

We present previous literature pertaining to the three tasks discussed in this
work. Additionally, we discuss subsequent extensions which combine convolutional and recurrent
networks to achieve improved results on activity recognition, image captioning,
and video description as well as related new tasks such as visual question answering.

\subsection{Prior Work}

\textbf{Activity Recognition.}
State-of-the-art shallow models combine spatio-temporal features along dense
trajectories \cite{wang2013action} and encode features as bags of words or
Fisher vectors for classification.
Such shallow features track how low level features change through time but cannot track higher level features.
Furthermore, by encoding features as bags of words or Fisher vectors,
temporal relationships are lost.

Many deep architectures proposed for activity recognition stack a fixed number of video frames for input to a deep network.
\cite{karpathy14cvpr} propose a fusion convolutional network which fuses layers which correspond to different input frames at various levels of a deep network.
\cite{simonyan2014two} proposes a two stream CNN which combines one CNN trained on RGB frames and one CNN trained on a stack of 10 flow frames.  
When combining RGB and flow by averaging softmax scores, results are comparable to state-of-the-art shallow models on UCF101\cite{soomro2012ucf101} and HMDB51\cite{kuehne2011hmdb}. 
Results are further improved by using an SVM to fuse RGB and flow as opposed to simply averaging scores.
Alternatively, \cite{Ji13tpami} and \cite{baccouche2011sequential} propose learning deep spatio-temporal features with 3D convolutional neural networks.  
\cite{baccouche2011sequential}, \cite{baccouche2010lstm} propose extracting visual and motion features and modeling temporal dependencies with recurrent networks.  
This architecture most closely resembles our proposed architecture for activity classification, though it differs in two key ways.  
First, we integrate 2D CNNs that can be pre-trained on large image datasets.  
Second, we combine the CNN and LSTM into a single model to enable end-to-end fine-tuning.

\textbf{Image Captioning.}
Several early works \cite{farhadi2010every,kulkarni2011baby, yang2011corpus,
mitchell2012midge} on image captioning combine object and scene recognition
with template or tree based approaches to generate captions.
Such sentences are typically simple and are easily distinguished from
more fluent human generated descriptions.  \cite{kuznetsova2012collective,
kuznetsova2014treetalk} address this by composing new sentences from existing
caption fragments which, though more human like, are not necessarily accurate or
correct.

More recently, a variety of deep and multi-modal
models \cite{frome2013devise,socher2014grounded,kiros2014multimodal,mrnn} have been proposed for
image and caption retrieval, as well as caption generation.
Though some of these models rely on deep convolutional nets for image
feature extraction~\cite{kiros2014multimodal,frome2013devise},
recently researchers have realized the importance of also including temporally
deep networks to model text.  \cite{socher2014grounded} propose an RNN to map sentences into a
multi-modal embedding space.  By mapping images and language into the same embedding space, 
they are able to compare images and descriptions for image and annotation
retrieval tasks.  \cite{mrnn} propose a model for caption generation
that is more similar to
the model proposed in this work: predictions for the next word are based on
previous words in a sentence and image features.  \cite{kiros2014multimodal} 
propose an encoder-decoder model for image caption retrieval
which relies on both a CNN and LSTM encoder to
learn an embedding of image-caption pairs. Their model uses a neural
language decoder to enable
sentence generation.  As evidenced by the rapid growth of image captioning,
visual sequence models like LRCN are increasingly important for
describing the visual world using natural language.

\textbf{Video Description.}
Recent approaches to describing video with natural language
have made use of templates, retrieval, or language models
\cite{guadarrama13iccv,rohrbach13iccv,khan11iccvw,barbu12uai,das13cvpr,
khan11iccvw,tan11mm,thomason14coling}.
To our knowledge, we present the first application of deep models to the video
description task.  Most similar to our work is \cite{rohrbach13iccv}, which use
phrase-based SMT \cite{koehn07acl} to generate a sentence. In Section
\ref{sec:videoDescription} we show that phrase-based SMT can be replaced with
LSTMs for video description as has been shown previously for language
translation \cite{sak14is,sutskever14nips}. 

\subsection{Contemporaneous and Subsequent Work}

Similar work in activity recognition and visual description was conducted contemporaneously with our work,
and a variety of subsequent work has combined convolutional and recurrent networks to both improve upon our results
and achieve exciting results on other sequential visual tasks.

\textbf{Activity Recognition.}
Contemporaneous with our work, \cite{ng2015beyond} train a network which combines CNNs and LSTMs for activity recognition.
Because activity recognition datasets like UCF101 are relatively small in comparison to image recognition datasets, \cite{ng2015beyond} pretrain their network using the Sports-1M \cite{karpathy14cvpr} dataset which includes over a million videos mined from YouTube.
By training a much larger network (four stacked LSTMs) and pretraining on a large video dataset, \cite{ng2015beyond} achieve 88.6\% on the UCF101 dataset.

\cite{yeung2015every} also combines a convolutional network with an LSTM to predict multiple activities per frame.
Unlike LRCN, \cite{yeung2015every} focuses on frame-level (rather than video-level) predictions, which allows their system to label multiple activities that occur in different temporal locations of a video clip. 
Like we show for activity recognition, \cite{yeung2015every} demonstrates that including temporal information improves upon a single frame baseline.
Additionally, \cite{yeung2015every} employ an attention mechanism to further improve results.

\textbf{Image Captioning.}
\cite{karpathy2014deep} and \cite{vinyals2014show} also propose models which combine a CNN with a recurrent network for image captioning.
Though similar to LRCN, the architectures proposed in \cite{karpathy2014deep} and  \cite{vinyals2014show} differ in how image features are input into the sequence model.
In contrast to our system, in which image features are input at each time step, \cite{karpathy2014deep} and \cite{vinyals2014show} only input image features at the first time step.
Furthermore, they do not explore a ``factored'' representation (Figure~\ref{fig:captionarchs}).
Subsequent work \cite{sat} has proposed attention to focus on which portion of the image is observed during sequence generation.
By including attention, \cite{sat} aim to visually focus on the current word generated by the model.
Other works aim to address specific limitations of captioning models based on combining convolutional and recurrent architectures.
For example, methods have been proposed to integrate new vocabulary with limited~\cite{mao2015learning} or no~\cite{hendricks2015deep} examples of images and corresponding captions. 

\textbf{Video Description.}
In this work, we rely on intermediate features for video description,
but end-to-end trainable models for visual captioning have since been proposed.
\cite{venugopalan2014translating} propose creating a video feature by pooling high level CNN features across frames.
The video feature is then used to generate descriptions in the same way an image is used to generate a description in LRCN.
Though achieving good results, by pooling CNN features, temporal information from the video is lost.
Consequently, \cite{venugopalan2015sequence} propose an LSTM to encode video frames into a fixed length vector before sentence generation with an LSTM.
Using an end-to-end trainable ``sequence-to-sequence'' model which can exploit temporal structure in video, \cite{venugopalan2015sequence} improve upon results for video description.
\cite{yao2015describing} propose a similar model, adding a temporal attention mechanism which weights video frames differently when generating each word in a sentence.

\textbf{Visual Grounding.}
\cite{rohrbach2015grounding} combine CNNs with LSTMs for visual grounding.
The model first encodes a phrase which describes part of an image using an LSTM, then learns to attend to the appropriate location in the image to accurately reconstruct the phrase.
In order to reconstruct the phrase, the model must learn to visually ground the input phrase to the appropriate location in the image.

\textbf{Natural Language Object Retrieval.}
In this work, we present methods for image retrieval based on a natural language description.
In contrast, \cite{hu2015natural} use a model based on LRCN for \textit{object retrieval}, which returns the bounding box around a given object as opposed to an entire image.
In order to adapt LRCN to the task of object retrieval, \cite{hu2015natural} include local convolutional features which are extracted from object proposals and the spatial configuration of object proposals in addition to a global image feature.
By including local features, \cite{hu2015natural} effectively adapt LRCN for object retrieval.

\section{Conclusion}
We've presented LRCN, a class of models that is both spatially and temporally deep, and flexible enough to be applied to a variety of vision tasks involving sequential inputs and outputs.
Our results consistently demonstrate that by learning sequential dynamics with a deep sequence model, we can improve upon previous methods which learn a deep hierarchy of parameters only in the visual domain, and on methods which take a fixed visual representation of the input and only learn the dynamics of the output sequence.

As the field of computer vision matures beyond tasks with static input and predictions, deep sequence modeling tools like LRCN are increasingly central to vision systems for problems with sequential structure.
The ease with which these tools can be incorporated into existing visual recognition pipelines makes them a natural choice 
for perceptual problems with time-varying visual input or sequential outputs, which these methods are able to handle with little input preprocessing and no hand-designed features.

\onecolumn 

\clearpage

\twocolumn

\ifCLASSOPTIONcompsoc

  \section*{Acknowledgments}

\else

  \section*{Acknowledgment}
\fi
The authors thank Oriol Vinyals for valuable advice and helpful discussion throughout this work.
This work was supported in part by DARPA's MSEE and SMISC programs, NSF awards IIS-1427425 and IIS-1212798, and the Berkeley Vision and Learning Center.
The GPUs used for this research were donated by NVIDIA.
Marcus Rohrbach was supported by a fellowship within the FITweltweit-Program of the German Academic Exchange Service (DAAD).
Lisa Anne Hendricks was supported by the NDSEG.

\ifCLASSOPTIONcaptionsoff
  \newpage
\fi

\bibliographystyle{IEEEtran}
\bibliography{IEEEabrv,refs}

\newenvironment{absolutelynopagebreak}
  {\par\nobreak\vfil\penalty0\vfilneg
   \vtop\bgroup}
  {\par\xdef\tpd{\the\prevdepth}\egroup
   \prevdepth=\tpd}

\begin{absolutelynopagebreak}
\vspace{-0.6in}

\begin{IEEEbiographynophoto}{Jeff Donahue}
  is a PhD student at the University of California, Berkeley, advised by Prof. Trevor Darrell.
  His research focuses on the use of deep learning for computer vision applications.
  He graduated with a BS in computer science from the University of Texas at Austin, where he was advised by Prof. Kristen Grauman.
\end{IEEEbiographynophoto}

\vspace{-0.20in}

\begin{IEEEbiographynophoto}{Lisa Anne Hendricks}
  is a PhD student at the University of California, Berkeley.
  Her research focuses on deep learning for sequential models as well as applications at the intersection of language and vision.
  She is advised by Prof. Trevor Darrell.
  Lisa Anne holds a Bachelor's of Science in Electrical Engineering (B.S.E.E.) from Rice University.
\end{IEEEbiographynophoto}

\vspace{-0.20in}

\begin{IEEEbiographynophoto}{Marcus Rohrbach\normalfont{'s}}
  research focuses on visual recognition, language understanding, and machine learning.
  He received his BSc and MSc degree in Computer Science from the University of Technology Darmstadt, Germany, in 2006 and 2009, respectively.
  From 2006-2007, he spent one year at the University of British Columbia as a graduate visiting student.
  During his PhD he worked at the Max Planck Institute for Informatics, Saarbr{\"u}cken, Germany with Bernt Schiele and Manfred Pinkal.
  He completed it in 2014 with summa cum laude at Saarland University and received the DAGM MVTec Dissertation Award 2015 for it.
  He currently works as a post-doc with Trevor Darrell at UC Berkeley.
\end{IEEEbiographynophoto}

\vspace{-0.20in}

\begin{IEEEbiographynophoto}{Subhashini Venugopalan}
  is a PhD student at the University of Texas at Austin.
  Her research focuses on deep learning techniques to generate descriptions for
  events in videos. She is advised by Prof. Raymond Mooney. 
  Subhashini holds a master's degree in Computer Science from IIT Madras and a
  bachelor's degree from NIT Karnataka, India.
\end{IEEEbiographynophoto}

\vspace{-0.20in}

\begin{IEEEbiographynophoto}{Sergio Guadarrama}
  is a Software Engineer at Google Research, where he works in Machine Perception as a member of the Vale team.
  He received his PhD from the Technical University of Madrid, followed by postdoctoral work at the European Center for Soft Computing.
  After that, he was first a Visiting Scholar and then a Research Scientist at UC Berkeley EECS. His research spans the areas of computer vision, language and deep learning.
  Dr. Guadarrama's current research focus is on new network architectures for multi-task dense predictions, such as object detection, instance segmentation, depth prediction and visual question-answering.
  He has received research grants from the Government of Spain, such as the Juan de la Cierva Award (Early Career Award in Computer Science), and the Mobility Grant for Postdoctoral Research.
\end{IEEEbiographynophoto}

\vspace{-0.20in}

\begin{IEEEbiographynophoto}{Kate Saenko}
  is an Assistant Professor of Computer Science at the University of Massachusetts Lowell, where she leads the Computer Vision and Learning Group.
  She received her PhD from MIT, followed by postdoctoral work at UC Berkeley EECS and Harvard SEAS.
  Her research spans the areas of computer vision, machine learning, and human-robot interfaces.
  Dr. Saenko's current research interests include domain adaptation of machine learning models and joint modeling of language and vision.
  She is the recipient of research grant awards from the National Science Foundation, DARPA, and other government and industry agencies.
\end{IEEEbiographynophoto}

\vspace{-0.20in}

\begin{IEEEbiographynophoto}{Trevor Darrell}
  is on the faculty of the CS Division of the EECS Department at UC Berkeley and is also appointed at the UCB-affiliated International Computer Science Institute (ICSI).
  He is the director of the Berkeley Vision and Learning Center (BVLC) and is the faculty director of the PATH center in the UCB Institute of Transportation Studies PATH.
  His interests include computer vision, machine learning, computer graphics, and perception-based human computer interfaces.
  Prof. Darrell received the SM and PhD degrees from MIT in 1992 and 1996, respectively.
  He was previously on the faculty of the MIT EECS department from 1999-2008, where he directed the Vision Interface Group.
  He was a member of the research staff at Interval Research Corporation from 1996-1999.
  He obtained the BSE degree from the University of Pennsylvania in 1988, having started his career in computer vision as an undergraduate researcher in Ruzena Bajcsy's GRASP lab.
\end{IEEEbiographynophoto}

\end{absolutelynopagebreak}

\end{document}